\documentclass[letter,oneside]{article}
\usepackage[margin=1.0in]{geometry}

\usepackage{amsmath,amsfonts,bm}

\def\eqref#1{equation~\ref{#1}}

\def\1{\bm{1}}

\DeclareMathAlphabet{\mathsfit}{\encodingdefault}{\sfdefault}{m}{sl}
\SetMathAlphabet{\mathsfit}{bold}{\encodingdefault}{\sfdefault}{bx}{n}

\DeclareMathOperator*{\argmax}{arg\,max}
\DeclareMathOperator*{\argmin}{arg\,min}

\usepackage{hyperref}
\usepackage{url}
\usepackage{xcolor}
\usepackage{microtype}
\usepackage{graphicx}
\usepackage{booktabs} 
\usepackage[utf8]{inputenc}
\usepackage{graphicx}
\usepackage{amsmath}
\usepackage{amsthm}
\usepackage{amssymb}

\newtheorem{proposition}{Proposition}

\usepackage{mdframed}
\usepackage{lipsum}
\usepackage{url}
\usepackage{caption}
\usepackage{multicol}
\usepackage{multirow}
\usepackage{wrapfig}
\usepackage{caption}
\usepackage{subcaption}
\usepackage[ruled, vlined]{algorithm2e}

\allowdisplaybreaks

\newcommand{\eat}[1]{}
\title{Local Explanations for Reinforcement Learning}

 \author{
 Ronny Luss\thanks{Equal contribution.}\\
   IBM Research, NY\\
      \texttt{rluss@us.ibm.com}
      \and
   Amit Dhurandhar$^*$\\
   IBM Research, NY\\
   \texttt{adhuran@us.ibm.com}
   \and
   Miao Liu\\
   IBM Research, NY\\
   \texttt{miao.liu1@ibm.com}
}

\date{}
\begin{document}

\maketitle

\begin{abstract}
Many works in explainable AI have focused on explaining black-box classification models. Explaining deep reinforcement learning (RL) policies in a manner that could be understood by domain users has received much less attention. In this paper, we propose a novel perspective to understanding RL policies based on identifying important states from automatically learned meta-states. The key conceptual difference between our approach and many previous ones is that we form meta-states based on locality governed by the expert policy dynamics rather than based on similarity of actions, and that we do not assume any particular knowledge of the underlying topology of the state space. Theoretically, we show that our algorithm to find meta-states converges and the objective that selects important states from each meta-state is submodular leading to efficient high quality greedy selection. Experiments on four domains (four rooms, door-key, minipacman, and pong) and a carefully conducted user study illustrate that our perspective leads to better understanding of the policy. We conjecture that this is a result of our meta-states being more intuitive in that the corresponding important states are strong indicators of tractable intermediate goals that are easier for humans to interpret and follow.
\end{abstract}

\section{Introduction}

Deep reinforcement learning (RL) has seen stupendous success over the last decade with superhuman performance in games such as Go \cite{alphago}, Chess \cite{alphazero}, and Atari benchmarks \cite{atari}. With increasing superior capabilities of automated (learning) systems, there is a strong push to understand the reasoning behind their decision making. One motivation is for (professional) humans to improve their performance in these games \cite{alphazeroLearn}. An even deeper reason is for humans to be able to trust these systems if they are deployed in real life scenarios \cite{xai}. 
The General Data Protection Regulation \cite{gdpr} passed in Europe demands that explanations need to be provided for any automated decisions that affect humans. While various methods have been provided to explain classification models \cite{lime,unifiedPI,LRPTOOLBOX,CEM} and be evaluated in an application-grounded manner \cite{rsi,tip}, the exploration of different perspectives to explain RL policies has been limited and user study evaluations \eat{comparing methods} are rarely employed in this space.

In this paper, we provide a novel perspective to produce human understandable explanations with a task-oriented user study that evaluates which explanations help users predict the behavior of a policy better. Our approach involves two steps: 1) learning meta-states, i.e., clusters of states, based on the dynamics of the policy being explained, and 2) within each meta-state, identifying states that act as intermediate goals, which we refer to as \emph{strategic states}. We call our method the Strategic State eXplanation
(SSX) method. Such an approach has real-world applicability. Consider scenarios where businesses want to increase their loyalty base. Companies often train RL policies to recommend next-best actions in terms of promotions to offer. They try to maximize the \emph{Lifetime Value} \cite{theo2020}. For such policies, SSX could identify strategic offers that lead to becoming loyalty customers based on a state space with features such as demographics, buying behavior, etc. Another example is robotics; consider the task of a robotic arm lifting a cup on a table which can be broken down to a sequence of stages, i.e., strategic states (subgoals) to aim for.

Contrary to the global nature of recent explainability works in RL \cite{apg, tldr, highlights}, our focus is on local explanations; given the current state, we explain the policy moving forward within a fixed distance from the current state. This key distinction lets us consider richer state spaces (i.e., with more features) because the locality restricts the size of the state space. 
It is also important to recognize the difference from bottlenecks \cite{qcut, relative_novelty} which are \emph{policy-independent} and learned by approximating the state space with randomly sampled trajectories; rather than help explain a policy, bottlenecks are used to \emph{learn} efficient policies such as through hierarchical RL \cite{botvinick2008} or options frameworks \cite{successoroptions}. Strategic states are rather learned with respect to a policy and identified without assuming access to the underlying topology.

\begin{figure*}[t]
\begin{center}
  \begin{tabular}{ccc}
   \includegraphics[width=.3\textwidth]{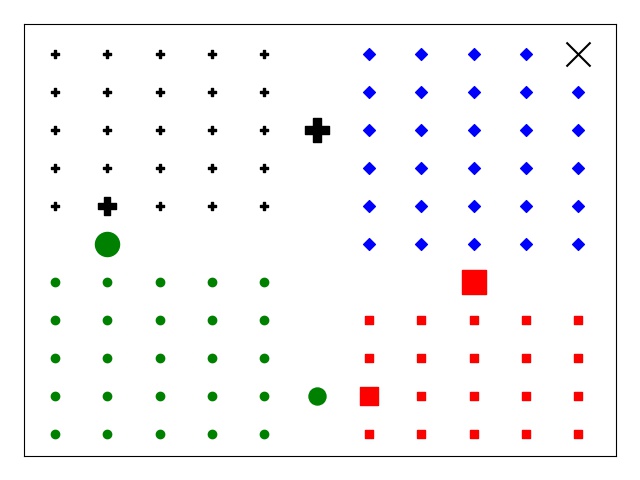} &
   \includegraphics[width=.3\textwidth]{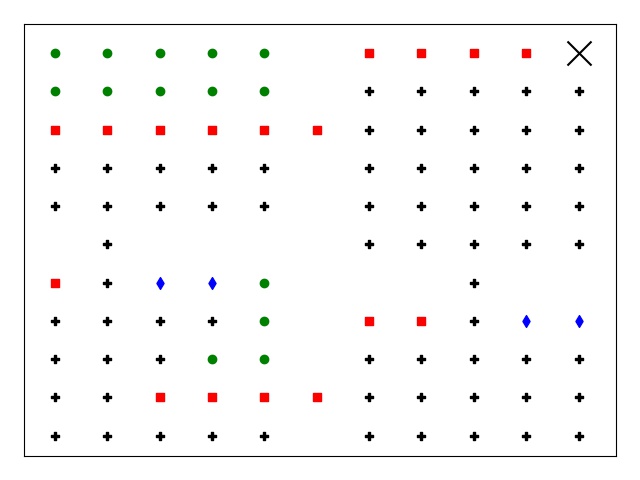} &
   \includegraphics[width=.3\textwidth]{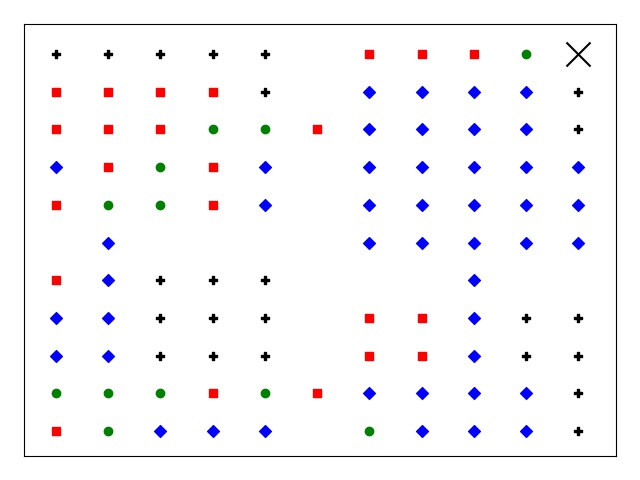} \\
    \small{(a)} & \small{(b)} & \small{(c) }
  \end{tabular}
  \caption{Illustrations of our SSX (a), VIPER (b), and abstract states used for compression (c) methods based on an expert policy for the Four Rooms game with neither having information about the underlying topology of the state space. Colors/Shapes denote different meta-states/clusters. The black \textbf{X} in the upper right is the goal state. \eat{Note that different colors/shapes in each visualization is solely for explaining the different methods, i.e., there is no relationship between the color/shapes used across (a), (b), and (c).} SSX clusters the four rooms exactly with strategic states denoted by larger markers, where the biggest marker implies the priority strategic state. SSX explains that the expert policy will head towards the open doors in each room preferring the door that leads to the room with the goal state.\eat{ Note the difference from bottlenecks which would find all doors but give no indication as to which doors to prioritize because bottlenecks are policy-independent as opposed to strategic states.}  VIPER clusters states by action (black/plus=up, green/circle=down, blue/diamond=left, red/square=right) based on the full (discrete) state space, rather than samples\eat{ as the original VIPER}, since it is tractable here. The compressed state space in (c) \eat{on which a policy can perform similarly to the expert policy trained on the original state space, and  also groups states as} is also a function of the experts (conditional) action distribution. Clusters in (b) and (c) are scattered making it challenging for a human to understand any policy over clusters.}
    \label{fig:four_rooms}
\end{center}
\vspace{-.4cm}
\end{figure*}

An example of this for the Four Rooms game is seen in Figure \ref{fig:four_rooms}a, where an agent moves through a grid with walls (represented by lack of a marker) looking for the goal state (upper right corner). Each position is a state and a meta-state is a cluster of possible positions (states sharing a color/marker). Within each meta-state, we identify certain states as \emph{strategic states} (shown with larger markers), which are intermediate states that moving towards will allow the agent to move to another meta-state and get closer to the goal state, which is the final state that the agent wants to get to. In Figure \ref{fig:four_rooms}a, each room is (roughly) identified as a meta-state by our method with the corresponding doors being the respective strategic states. Topology refers to the graph connecting states to one another; our method only has access to the knowledge of which states are connected (through the policy), whereas reinforcement learning algorithms might have access to properties of the topology, e.g., the ability to access similar states using successor representations \cite{eigenoptions}. In Figure \ref{fig:four_rooms}, the topology is a graph connecting the different positions in each room or the doors connecting two rooms.

A key conceptual difference between our approach and others \eat{global (and even local)explainable RL methods }is that other methods aggregate insight (i.e. reduce dimension) as a function of actions \cite{viper} or formulas derived over factors of the state space \cite{tldr} to output a policy summary, whereas we aggregate based on locality of the states determined by the expert policy dynamics and further identify strategic states based on these dynamics. Other summarization methods simply output simulated trajectories deemed important \cite{highlights,highlights-sal} as judged by whether or not the action taken at some state matters. We use the term \emph{policy dynamics} to refer to state transitions and high probability paths. We use the term dynamics because this notion contrasts other methods that use actions to explain what to do in a state or to identify important states; strategic states are selected according to the trajectories that lead to them, and these trajectories are implicitly determined by the policy.\eat{Note that locality is \emph{not} determined assuming knowledge of the underlying structure or topology of the state space.} 

The example in Figure \ref{fig:four_rooms} exposes the global view of our explanations when the state space is small because local approximations of the state space are not needed. We show that this perspective leads to more understandable explanations; aggregating based on actions, while precise, are too granular a view where the popular idiom \emph{can't see the forest for the trees} comes to mind. We conjecture that the improved understanding is due to our grouping of states being more intuitive with strategic states indicating tractable intermediate goals that are easier to follow. An example of this is again seen in Figures \ref{fig:four_rooms}b and \ref{fig:four_rooms}c, where grouping based on actions for interpretability or for efficiency leads to less intuitive results (note that Figure \ref{fig:four_rooms}c replicates Figure 4b from \cite{stateabs}). This scenario is further discussed in Section \ref{sec:exp}, where yet other domains have large state spaces and require strategic states to explain local scenarios. 
As such, our main contributions are two-fold:\\
1. We offer a novel framework for understanding RL policies, which to the best of our knowledge, differs greatly from other methods in this space which create explanations based on similarity of actions rather than policy dynamics. We demonstrate on four domains of increasing difficulty.\\
2. We conduct a task-oriented user study to evaluate effectiveness of our method. Task-oriented evaluations are one of the most thorough ways of evaluating explanation methods \cite{rsi,lipton2016mythos,tip} as they assess simulatability
, yet to our knowledge, have rarely been used in the RL space.

\section{Related Work}
\label{sec:relw}
While a plethora of methods are proposed in XAI \cite{lime,unifiedPI,LRPTOOLBOX,CEM}, we focus on works related to RL explainability and state abstraction, as they are most relevant to our work, and distinguish between global and local explainability methods as done in \cite{burkart2021}. Namely, global methods are \emph{model explanation approaches} whereas local methods are \emph{instance explanation approaches}. It is important to note that various global methods described below, e.g. decision trees such as \cite{viper}, can be used to explain individual instances, however this does not apply to all global methods, e.g. \cite{highlights}. While global methods can explain a model without passing individual instances, i.e., by analyzing the splits of a decision tree, local methods only explain a model's performance on individual instances.

In the spirit of \cite{burkart2021}, most global RL methods summarize a policy using some variation of state abstraction where the explanation uses aggregated state variables that group actions \cite{viper, liu2021} using decision trees or state features \cite{apg} using importance measures, or such that an ordering of formulas based on features is adhered to \cite{tldr}. These approaches all intend to provide a global summary of the policy. \cite{liu2021} is most recent and can be viewed complementary as well; the idea of using latent representations to increase interpretability could be adapted in our framework when visualizing results. Other summaries output trajectories deemed important according to importance measures \cite{highlights, highlights-sal} or through imitation learning \cite{lage2019}, or train finite state representations to summarize a policy with an explainable model \cite{rpn_learning, rpn_reunderstanding}. Visualization techniques combined with saliency have been used to either aggregate states and view the policy from a different perspective \cite{graying_black_box} or create a trajectory of saliency maps \cite{greydanus2018}. Other works try to find state abstractions or simplify the policy \cite{stateabs,subgoalim,linpolicy}, which should not be confused with works seeking explainability. State abstraction in these works is used for compression so that simpler policies can be used; the compressed state space is not interpretable as seen in Figure \ref{fig:four_rooms}c.

Turning towards local explanation methods, some works focus on self-explaining models \cite{softattneX} where the policy has soft attention and so can indicate which (local) factors it is basing its decision on at different points in the state space. \cite{yau2020} learns a \emph{belief map} concurrently during training which is used to explain locally by predicting the future trajectory. Interestingly, there are works which suggest that attention mechanisms should not be considered as explanations \cite{attnnotex}. These directions focus on learning an inherently explainable model rather than explaining a given model. Other works use local explanation methods to explain reasons for a certain action in a particular state \cite{cemrl,actinfl}. These are primarily contrastive where side information such as access to the causal graph may be assumed. Our approach is both methodologically and conceptually different.\eat{, where we form meta-states based on policy dynamics and then identify (strategic) states through which many policy-driven paths cross.}

There are also program synthesis-type methods \cite{reprogrl,progsynth} that learn syntactical programs representing policies, which while more structured in their form, are typically not amenable to lay users. Methods in safe RL try to uncover failure points of a policy \cite{criticalstate} by generating critical states. Another use of critical states\eat{, defined differently by how actions affect the value of a state,} is to establish trust in a system \cite{criticalstate_trust}. There is also explainability work in the Markov decision processes literature focusing on filling templates according to different criteria such as frequency of state occurrences or domain knowledge \cite{mdp_sufficient, mdp_explanations}. An elaborate discussion of these and other methods can be found in \cite{surveyexrl}, all of which unequivocally are different from ours.

\eat{
\begin{algorithm}[ht]
\SetAlgoLined

\textbf{Input:} State space $\mathcal{S}$, action space $\mathcal{A}$, expert policy $\pi_E(\cdot,\cdot) : (\mathcal{A},\mathcal{S}) \rightarrow \mathbb{R}$, number of meta-states $k$, convergence threshold for finding meta-states $\epsilon_{\phi}>0$, convergence threshold for finding strategic states $\epsilon_{g}>0$ and out-paths regularizer $\eta>0$
\vspace{3mm}

1) Compute maximum likelihood path matrix $\Gamma$

2) Find meta-states $\mathcal{S}_\phi=\mathsf{MS}(\mathcal{S},\mathcal{A},\pi_E,\Gamma,k,\epsilon_{\phi},\eta)$

3) Find strategic states $G_{\Phi}=\mathsf{SS}(\mathcal{S}_\phi,\Gamma,\epsilon_{g})$

\KwOut{Meta-states $\mathcal{S}_\phi$ and strategic states $G_{\Phi}$ corresponding to each meta-state. }
 \caption{Strategic State eXplanation (SSX) method.}
  \label{algo:SSX}
\end{algorithm}
}
\section{Method}
\label{sec:meth}
We now describe our algorithm, the Strategic State eXplanation (SSX) method, which involves computing shortest paths between states, identifying meta-states, and selecting their corresponding strategic states. 
Recall that all paths discussed below are based on transitions dictated by an expert policy we want to explain; bottlenecks however, are identified from paths generated as random walks through the state space and are meant to help \emph{learn} policies rather than explain them.

\noindent\textbf{Notations:} Let $\mathcal{S}$ define the full state space and $s\in\mathcal{S}$ be a state in the full state space.  Denote the expert policy by $\pi_E(\cdot,\cdot) : (\mathcal{A},\mathcal{S}) \rightarrow \mathbb{R}$ where $\mathcal{A}$ is the action space. The notation $\pi_E \in \mathbb{R}^{|\mathcal{A}| \times |\mathcal{S}|}$ is a matrix where each column is a distribution of actions to take given a state (i.e., the policy is stochastic). We assume a transition function $f_E(\cdot, \cdot) : (\mathcal{S},\mathcal{S}) \rightarrow \mathbb{R}$ that defines the likelihood of moving between states in one jump by following the expert policy. Let $\mathcal{S}_\phi=\{\Phi_1,...,\Phi_k\}$ denote a meta-state space of cardinality $k$\eat{ $\phi(\cdot) : \mathcal{S} \rightarrow \mathcal{S}_\phi$ denote a meta-state mapping such that $\phi(s)\in\mathcal{S}_\phi$ is the meta-state assigned to $s\in\mathcal{S}$}. Denote $m$ strategic states of meta-state $\Phi$ by $G^{\Phi}=\{g^{\Phi}_1,...,g^{\Phi}_m\}$ where $g^{\Phi}_i\in \mathcal{S}$ $\forall i\in \{1,...,m\}$.

\noindent\textbf{Maximum likelihood (expert) paths:} One criterion used below is that two states in the same meta-state should not be far  from each other. The distance we consider is the most likely path from state $s$ to state $s'$ under $\pi_E$.  Consider a fully connected, directed \eat{(in both directions)} graph where the states are vertices and an edge from $s$ to $s'$ has weight $-\log{f_E(s, s')}$. By this definition, the shortest path is also the maximum likelihood path from $s$ to $s'$. Denote by $\gamma(s, s')$ the value of this maximum likelihood path and $\Gamma\in \mathbb{R}^{|\mathcal{S}|\times |\mathcal{S}|}$ a matrix containing the values of these paths for all pairs of states in the state space. 

\eat{We now want to group states and form meta-states such that the sum of maximum likelihoods across all pairs of states within each meta-state $\Phi_i$, i.e. $\sum_{s,s'\in\Phi_i}{\gamma(s,s')}$, is maximized. This however, is just the first criterion, we now describe are second criterion.}

\noindent\textbf{Counts of Out-paths:} Another criterion used below for assigning states to meta-states is that if state $s$ lies on many of the paths between one meta-state $\Phi_i$ and all other meta-states, then $s$ should be assigned the meta-state $\Phi_i$, i.e., $s\in\Phi_i$. We define below the number of shortest paths leaving $\Phi_i$ that a fixed state $s$ lies on. Denote $T(s,s')$ as the set of states that lie on the maximum likelihood path between $s$ and $s'$, i.e., the set of states that define $\gamma(s, s')$. Then $1[s \in T(s', s'')]$ is the indicator of whether state $s$ lies on the maximum likelihood path between $s'$ and $s''$, and we compute the count of the number of such paths for state $s$ and \eat{its assigned }meta-state $\Phi$ via
\eat{\begin{equation}
\label{eq:pathcount}
    C(s, \phi(s)) = \sum_{\substack{s'\ne s,\\ \phi(s')=\phi(s)}}\sum_{\substack{s'' :\\ \phi(s'') \ne \phi(s)}} 1[s \in T(s', s'')].
\end{equation}}
\begin{equation}
\label{eq:pathcount}
    C(s, \Phi) = \sum_{\substack{s'\ne s, s'\in\Phi}}\hspace{0.5em}\sum_{ s'' \notin \Phi} 1[s \in T(s', s'')].
\end{equation}
\eat{where our regularization here is to choose a meta-state $\phi(s)$ that maximizes $C(s, \phi(s))$. We desire this so that potentially high quality strategic states will belong to the appropriate meta-state.}
One may also consider the likelihood (rather than count) of out-paths by replacing the indicator in eq. (\ref{eq:pathcount}) with $\gamma(s',s'')$. $C(s,\phi(s))$ can be computed for all $s\in\mathcal{S}$ in $O(|\mathcal{S}|^2)$ by iteratively checking if predecessors of shortest paths from each node to every other node lie in the same meta-state as the first node on the path. Approximating $C(s,\phi(s))$ (through sampling) can lead to significant computational savings while maintaining stability of the selected strategic states. The computation of out-paths in equation (\ref{eq:pathcount}) involves searching over all paths between states in each meta-state with those states in other meta-states. See Appendix B where stability is illustrated when randomly sampling a fixed fraction of the states in other meta-states (second summation in equation (\ref{eq:pathcount})).

\subsection{Learning Meta-States}
We seek to learn meta-states that balance the criteria of having high likelihood paths within the meta-state and having many out-paths from states within the meta-state. It is important to distinguish our goals from more classic cluster methods that are solely state-based; such clusterings would be independent of the expert policy that we want to explain and hence could lead to states connected by low likelihood paths as per the expert policy being in the same meta-state. Our meta-states account for the expert policy by minimizing the following objective for a suitable representation of $s$, which in our case is the eigen-decomposition of the Laplacian of $\Gamma$:
\begin{equation}
\label{eq:mainobj}
    \argmin\limits_{\mathcal{S}_{\phi}}\sum_{\Phi \in \mathcal{S}_{\phi}}\sum_{s\in \Phi} \left[(s-c_{\Phi})^2-\eta C(s, \Phi)\right]
\end{equation}
where $c_{\Phi}$ denotes the centroid of the meta-state $\Phi$ and $\eta>0$ balances the trade-off between the criteria. Note that we are optimizing $\mathcal{S}_{\phi}$ over all possible sets of meta-states. Other representations for $s$ and functions for the first term could be used; our choice is motivated from the fact that such formulations are nostalgic of spectral clustering \cite{spectral} which is known to partition by identifying well-connected components, something we strongly desire. This representation connects the explanation to the policy because the matrix $\Gamma$ is determined by the policy and provides intuitions. Specifically, in problem (2) when $\eta\rightarrow 0$, the meta-states will tend to be equi-sized where the likelihood of meta-state transitions will be minimized leading to (approximate) optimization of an NCut objective \cite{spectral_clustering_tutorial}. For larger $\eta$, the likelihood of meta-state transitions is still kept small (which is desirable), with a tendency towards having a few large meta-states. We found our method to be stable for $\eta\in (0,5]$.

Our method for solving eq. (\ref{eq:mainobj}) is given by algorithm \ref{algo:meta-state} and can be viewed as a regularized version of spectral clustering. In addition to clustering a state with others that it is connected to, the regularization pushes a state to a cluster, even if there are only a few connections to the cluster, if the policy dictates that many paths starting in the cluster go through that state.

\begin{algorithm}[h]
\SetAlgoLined
1) Get eigen representation of each state $s$ from eigen decomposition of the Laplacian of $\Gamma$

2) Randomly assign states $s\in \mathcal{S}$ to a meta-state in $\mathcal{S}_\phi=\{\Phi_1,...,\Phi_k\}$ and compute  centroids $c_1,...,c_k$ for meta-states

3) $\xi^{\text{cur}}=$ current value of objective in eq. (\ref{eq:mainobj})

\SetKwRepeat{Do}{do}{while}
\Do{$|\xi^{\text{cur}}-\xi^{\text{prev}}| \ge \epsilon_{\phi}$}
 {
 4) $\xi^{\text{prev}}=\xi^{\text{cur}}$
 
 5) Reassign states $s$ to the meta-states based on smallest value of $(s-c_{\Phi})^2-\eta C(s, \Phi)$
 
 6) Compute centroids $c_1,...,c_k$ for meta-states based on current assignment
 
 7) $\xi^{\text{cur}}=$ current value of objective in eq. (\ref{eq:mainobj})
 }
 
 \KwOut{Meta-states $\{\Phi_1,...,\Phi_k\}$  }
 \caption{Meta-states $\mathsf{MS}(\mathcal{S},\mathcal{A},\pi_E,\Gamma,k,\epsilon_{\phi},\eta)$}
  \label{algo:meta-state}
\end{algorithm}

 \begin{algorithm}[h]
\SetAlgoLined
 
  \For{$i=1$ to $k-1$} 
 {
 1) Let $\xi^{\text{cur}}=0$ and $G_{\Phi_i}=\emptyset$
 
\eat{ 2) Let $G_{\Phi_i}=\emptyset$}
 
 \SetKwRepeat{Do}{do}{while}
\Do{$|\xi^{\text{cur}}-\xi^{\text{prev}}| \ge \epsilon_{g}$}
 {
 2) $\xi^{\text{prev}}=\xi^{\text{cur}}$
 
 \eat{4) Let $g = \argmax\limits_ {s\in \Phi_i\setminus G_{\Phi_i}}$ of eq. (\ref{eq:strategic}) given the current set of strategic states $G_{\Phi_i}$
 }
 3) $G_{\Phi_i}=G_{\Phi_i}\cup g$ where $g$ solves eq. (3) over states not in the set of strategic states $G_{\Phi_i}$
 
 4) $\xi^{\text{cur}}=$ evaluate eq. (\ref{eq:strategic}) with $G_{\Phi_i}$
 
 }
 }
 5) $G_{\Phi_k}=\mathsf{g}$, where $\mathsf{g}$ is the expert policy's goal state
 
 \KwOut{Strategic states for each corresponding meta-state $\{G_{\Phi_1},...,G_{\Phi_{k}}\}$ }
 \caption{Strategic State function $\mathsf{SS}(\mathcal{S}_\phi,\Gamma,\epsilon_{g})$. Finds Strategic States with Greedy Selection (w.l.o.g. assume meta-state $\Phi_k$ contains the goal state).}
  \label{algo:SS}
\end{algorithm}

\subsection{Identifying Strategic States}

Next, strategic states are selected for each meta-state. Assume that $g^{\Phi}_1,...,g^{\Phi}_m\in \mathcal{S}$ are $m$ strategic states for a meta-state $\Phi$ that does not contain the target state. SSX finds strategic states by solving the following problem for some $\lambda>0$: 
\begin{align}
\label{eq:strategic}
    G^{(m)}_{\Phi}=\argmax\limits_{g^{\Phi}_1,...,g^{\Phi}_m} \sum_{i=1}^m C(g^{\Phi}_i, \Phi)-\lambda\sum_{i=1}^{m-1}\sum_{j= i+1}^m \max\left( \gamma(g^{\Phi}_i,g^{\Phi}_j),\gamma(g^{\Phi}_j,g^{\Phi}_i)\right).
\end{align}

The first term favors states that lie on many out-paths from the meta-state, while the second term favors states that are far from each other. The overall objective tries to pick states that explore different meta-states consistent with the expert policy, while balancing the selection of states to be diverse \eat{(i.e., far from each other)}. The objective in eq. (\ref{eq:strategic}) is submodular as stated next (proof in Appendix A) and hence we employ greedy selection in algorithm \ref{algo:SS}. Note that for the meta-state that contains the target state, the target state itself is its only strategic state.

\begin{proposition}
The objective to find strategic states in equation (\ref{eq:strategic}) is submodular.
\end{proposition}

\subsection{Strategic State eXplanation (SSX) method} \label{ss:ssx}
Our method is detailed as follows. First, the maximum likelihood path matrix $\Gamma$ is computed. Then, algorithm \ref{algo:meta-state} tries to find meta-states that are coherent w.r.t. the expert policy, in the sense that we group states into a meta-state if there is a high likelihood path between them. If many paths from states in a meta-state go through another state, then the state is biased to belong to this meta-state. Finally, algorithm \ref{algo:SS} selects strategic states by optimizing a trade-off between being on many out-paths with having a diverse set of strategic states.
\subsection{Scalability and Complexity}
\label{ss-practical_considerations}
Given our general method, we now discuss important details for making our algorithm practical when applied to different domains. SSX is applied in Section \ref{sec:exp} to games with state spaces ranging from small to exponential in size. SSX is straightforward for small state spaces as one can pass the full state space as input, however, neither finding meta-states nor strategic states would be tractable with an exponential state space. One approach could be to compress the state space using VAEs as in \cite{stateabs}, but as shown in Figure \ref{fig:four_rooms}c, interpretability of the state space can be lost as there is little control as to how states are grouped. 
Our approach is to use local approximations to the state space; given a starting position, SSX approximates the state space by the set of states within some $N>0$ number of moves from the starting position. In this approach, Algorithms \ref{algo:meta-state} and \ref{algo:SS} are a function of $N$, i.e., increasing $N$ increases the size of the approximate state space which is passed to both algorithms. One can contrast our approach of locally approximating the state space with that of VIPER \cite{viper} which uses full sample paths to train decision trees. While the number of states in such an approximation is $M^N$, where $M$ is the number of possible agent actions, the actual number of states in a game such a pacman is much smaller in practice. Indeed, while pacman has 5 possible actions, growth of the state space in our approximation as $N$ increases acts similar to a game with 2-3 actions per move because most states in the local approximation are duplicates due to both minipacman and the ghost going back and forth. See Figure 5 in Appendix B, where other practical considerations, including approximating $C(s, \Phi)$, tractability of $\Gamma$ and the eigen decomposition of its Laplacian, are also discussed.

\eat{\noindent\textbf{Number of Meta-states $k$:} The number of meta-states can be chosen using standard techniques as trying different $k$ and finding the knee of the objective (i.e. where the objective has little improvement) or based on domain knowledge. State representations may affect the (appropriate) number. }


\section{Experiments}
\label{sec:exp}
This section illustrates the Strategic State eXplanation (SSX) method on three domains: four rooms, door-key, and minipacman. These domains represent different reinforcement learning (RL) regimes, namely, 1) non-adversarial RL with a small state space and tabular representation for the policy, 2) non-adversarial RL, and 3) adversarial RL, the latter two both with a large state space and a deep neural network for the policy. These examples illustrate how strategic states can aid in understanding RL policies. A fourth domain, pong, represents adversarial RL where the environment has no access to the adversary and is in Appendix C. Lack of access to the adversary means that the maximum likelihood path matrix $\Gamma$ requires simulation.  Experiments were performed with 1 GPU and up to 16 GB RAM. The number of strategic states was chosen such that additional strategic states increased the objective value by at least 10\%. The number of meta-states was selected as would be done in practice, through cross-validation to satisfy human understanding. Experiments demonstrating stability of strategic states to changes in the initial state, i.e. robustness of SSX to the initial state, as well as how sensitive strategic states are to the size of the local approximation, using measures of stability and faithfulness, are in Appendix E. Details about environments are in Appendix F.\\

\begin{figure*}[t]
\begin{center}
  \begin{tabular}{cccc|cccc}
    \multicolumn{4}{c|}{Locked Door} &\multicolumn{4}{c}{Unlocked Door}\\ 
   \includegraphics[width=.09\textwidth]{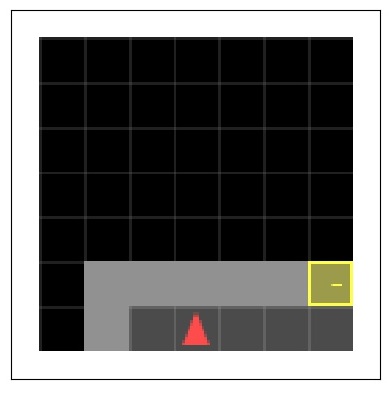} &
   \includegraphics[width=.09\textwidth]{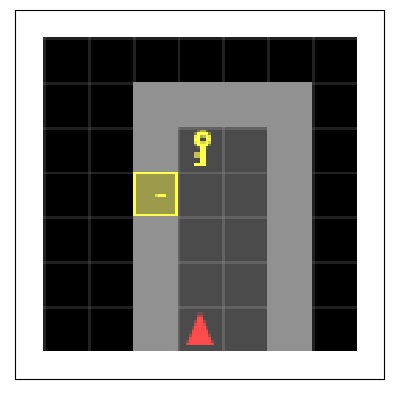} &
   \includegraphics[width=.09\textwidth]{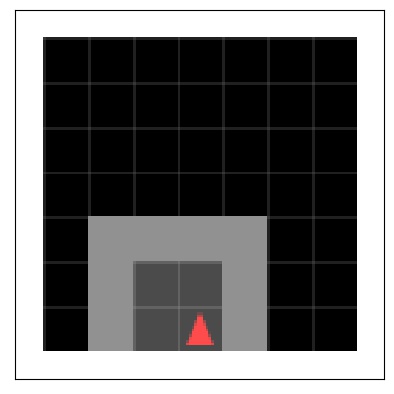} &
   \includegraphics[width=.09\textwidth]{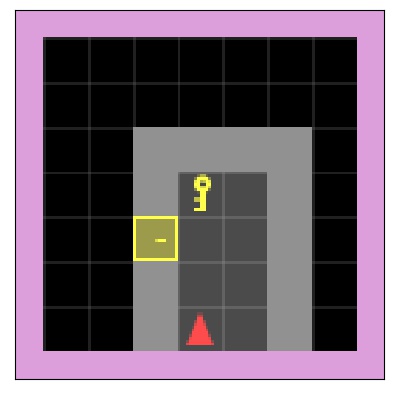} &
   \includegraphics[width=.09\textwidth]{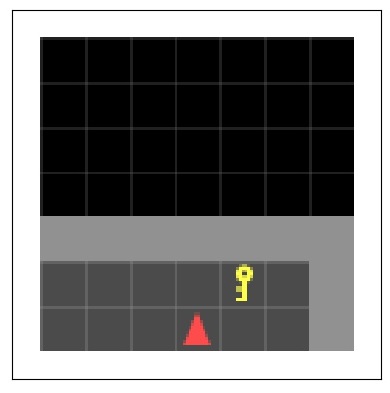} &
   \includegraphics[width=.09\textwidth]{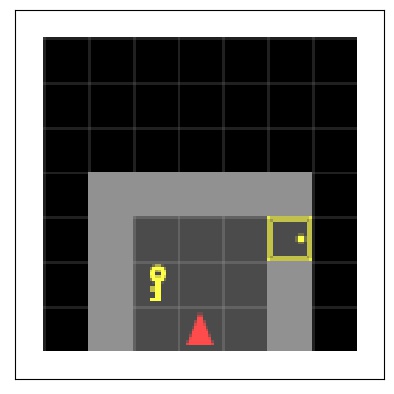} &
   \includegraphics[width=.09\textwidth]{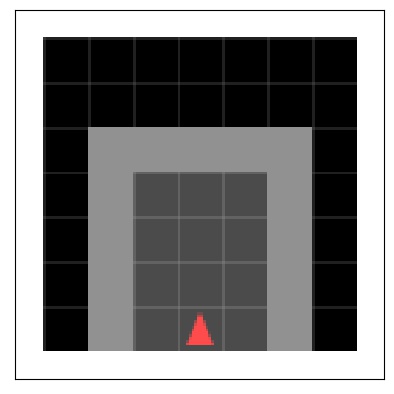} &
   \includegraphics[width=.09\textwidth]{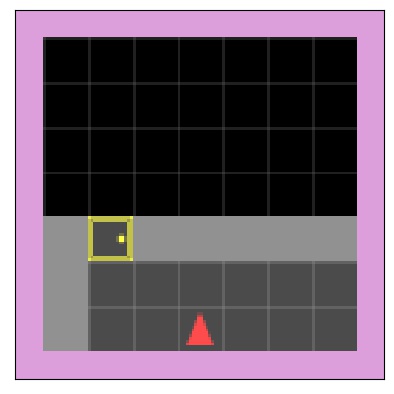} \\
   
   \includegraphics[width=.09\textwidth]{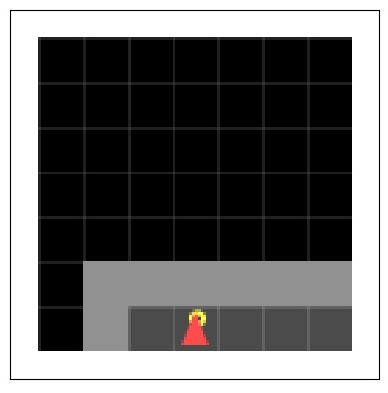} &
   \includegraphics[width=.09\textwidth]{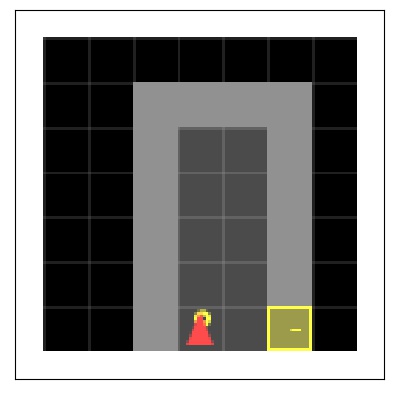} &
   \includegraphics[width=.09\textwidth]{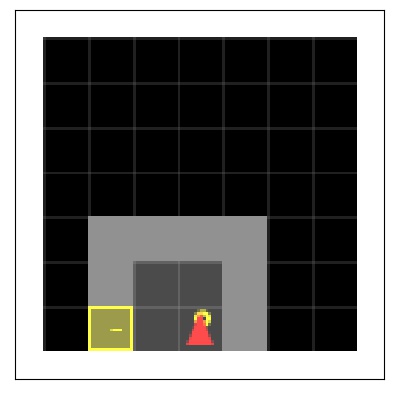} &
   \includegraphics[width=.09\textwidth]{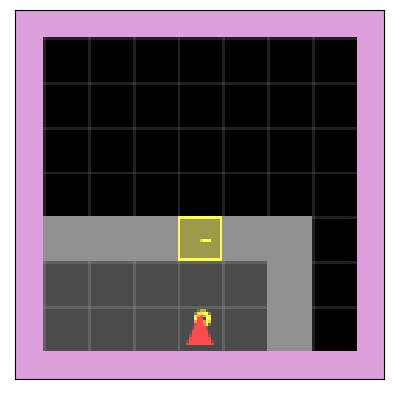} &
   \includegraphics[width=.09\textwidth]{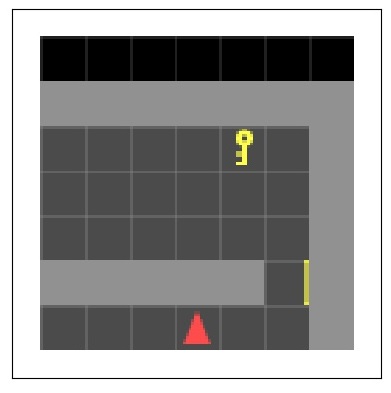} &
   \includegraphics[width=.09\textwidth]{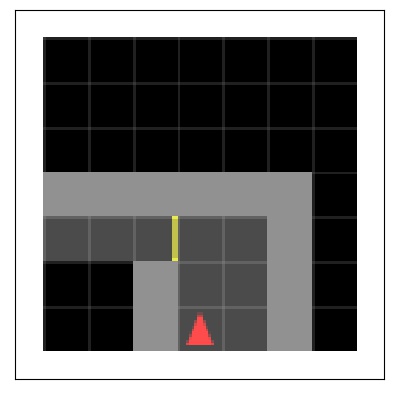} &
   \includegraphics[width=.09\textwidth]{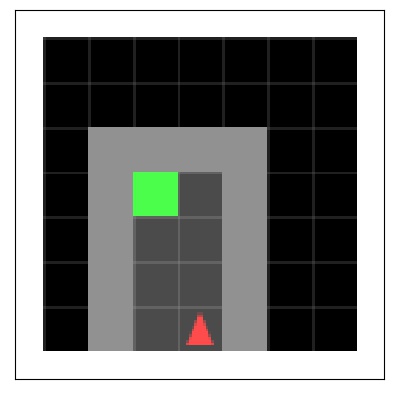} &
   \includegraphics[width=.09\textwidth]{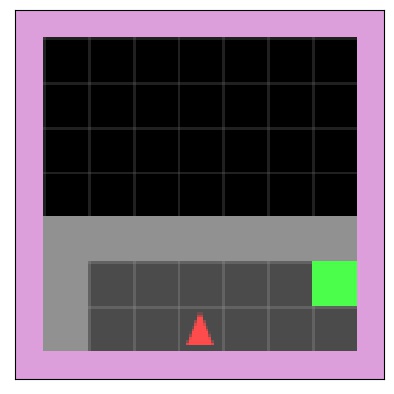}
   
\eat{   \includegraphics[width=.09\textwidth]{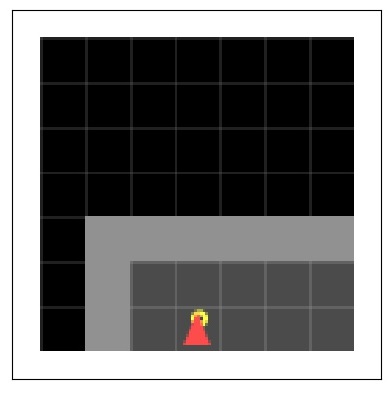} &
   \includegraphics[width=.09\textwidth]{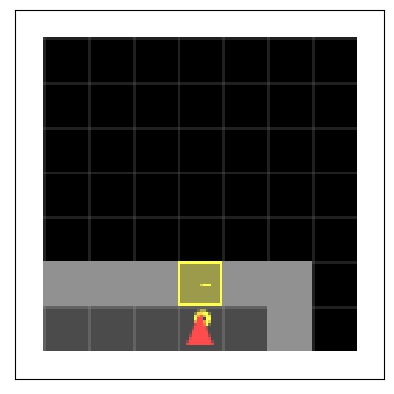} &
   \includegraphics[width=.09\textwidth]{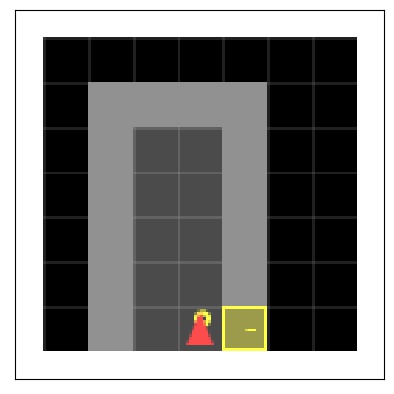} &
   \includegraphics[width=.09\textwidth]{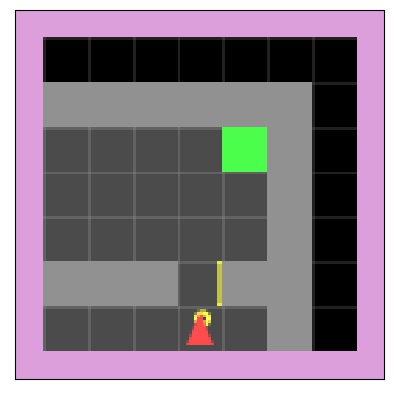} &
   \includegraphics[width=.09\textwidth]{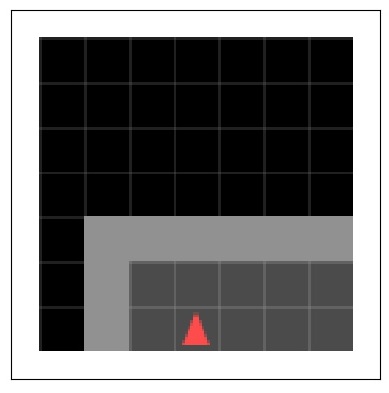} &
   \includegraphics[width=.09\textwidth]{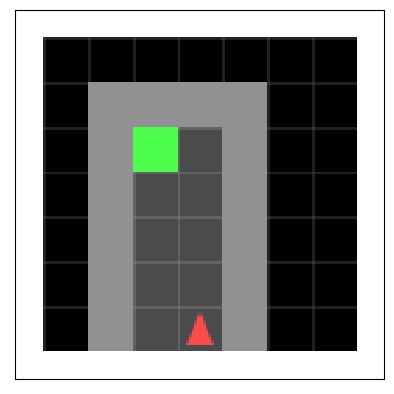} &
   \includegraphics[width=.09\textwidth]{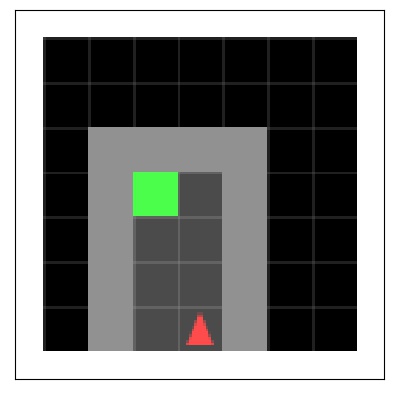} &
   \includegraphics[width=.09\textwidth]{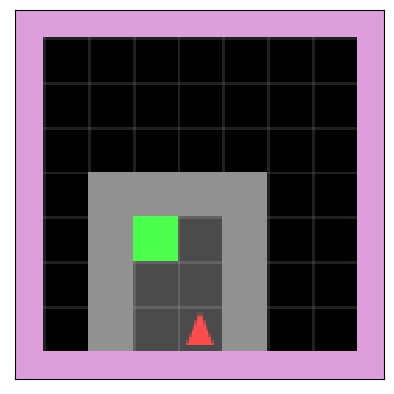}} 
 
   \end{tabular}
     \caption{Illustration of our SSX method on Door-Key. Policies were trained on two different environments: Locked Door and Unlocked Door.  Each row corresponds to a meta-state and strategic state (outlined in pink) from running SSX starting at a different number of moves into the same path (one path for completing the task in each of the two environments).\eat{ For the Locked Door environment, the agent looks for the key (row 1) and then gets the key (row 2). For the Unlocked Door environment, the agent looks for the door (row 1) and then opens and goes through the door (row 2).}}
    \label{fig:doorkey_strategic_states}
  \end{center}
  \vspace{-0.5cm}
 \end{figure*}
\noindent\textbf{Four Rooms:} The objective of Four Rooms is to move through a grid and get \eat{from the initial state (lower left  corner) }to the goal state (upper right corner).\eat{ A player can move left, right, up, or down. Players can only move to a position that has a color marker.} The lack of a marker in a position represents a wall. Our grid size is $11\times 11$.\eat{ that uses the framework from \url{https://github.com/david-abel/rl_info_theory} \cite{abide1}} The state space consists of the player's current position and the policy is learned as a tabular representation, since the state space is not large, using Value Iteration \cite{abide1}. 

SSX is displayed in Figure \ref{fig:four_rooms}a with settings that learn four meta-states. Clustering the states using algorithm \ref{algo:meta-state} according to the policy dynamics (i.e. maximum likelihood path matrix $\Gamma$) results in an (almost) perfect clustering of states according to the rooms. Larger markers denote strategic states learned in each meta-state, with the larger strategic state in each room corresponding to the first strategic state found. Clearly either door in each room could lead to the goal state in the upper right corner (black X), but it is important to note that higher valued strategic states in the red and black rooms are those that lead directly to the blue room containing the goal state.

Figure \ref{fig:four_rooms}b illustrates the results of VIPER \cite{viper}.\eat{ In this case we apply VIPER-D to the entire state space as sampling of paths is not required given that the state space is not too large.} The explanation is illustrated using different colors per action which effectively offers decision tree rules. While an explanation based on rules can be informative in continuous state spaces (as demonstrated in \cite{viper}), such rules applied to a discrete state space as done here may lead to confusion, e.g., groups of reds states are split by black states in the lower left room and allow for an optimal policy but it is not clear how to describe the cluster of states in which to take each action.\eat{ As is demonstrated via a user study in Section \ref{sec:user_study} on minipacman, the visualization of strategic states is much more clearly understandable to a user than this form of grouping.} Figure \ref{fig:four_rooms}c illustrates the difference between explainability and compression\eat{ when considering meta-states} \cite{stateabs} where one wants to learn abstract states upon which a proxy policy replicating the expert policy can be efficiently learned on the full state space. The lack of interpretability of the abstract states is not of concern in that context.\\

\noindent\textbf{Door-Key:} Door-Key is another non-adversarial game, but differs from Four Rooms because the state space is exponential in the size of the board. The policy is learned as a convolutional neural network with three convolutional and two linear layers.\eat{ following the training framework in \url{https://github.com/lcswillems/rl-starter-files} which uses the Door-Key environment in \url{https://github.com/maximecb/gym-minigrid}.} In this game, one must navigate from one room through a door to the next room and find the goal location to get a reward. Policies are trained under two scenarios. In the first scenario, a key in the first room must be picked up and used to unlock the door before passing through. In the second scenario, the door is closed but unlocked, so one does not need to first pick up the key to open the door.

SSX is run with local approximations to the state space with the maximum number of steps set to 6 as discussed in Section \ref{ss-practical_considerations}. Results are shown in Figure \ref{fig:doorkey_strategic_states}. The state space is a $7\times7$ grid reflecting the forward facing perspective of the agent. Walls are light gray and empty space visible to the agent is dark gray. Grid positions blocked from view by walls are black. The scenes in Figure \ref{fig:doorkey_strategic_states} are exactly what a user sees. To better understand why scenes do not appear easily connected, consider the first two states in the first row - the only difference \eat{from the first state }is that the agent changed directions. When facing the wall, the agent's view only includes the three positions to the right and one position to the left. Positions on the other side of the wall are not visible to the agent, which is depicted as black. When the agent changed directions\eat{ (row 1, column 2)}, many more positions in the room become visible to the agent. \eat{Locked, unlocked, and open doors are represented by yellow squares shaded over dark gray, dark gray squares outlined by yellow, and dark gray squares with one side yellow, respectively.} 

In Figure \ref{fig:doorkey_strategic_states}, a sample path was generated using each policy. SSX was run at three different states along these paths, and one meta-state and corresponding strategic state (outlined in pink) from each SSX explanation is displayed.\eat{ Explainability provided by SSX is used to distinguish between the locked and unlocked door policies; given a sample path to solve each task, SSX was run at different states along the path, three of which are shown for each environment with one meta-state and corresponding strategic state (outlined in pink) displayed.} The two strategic states for the locked door environment correspond to the agent looking for the key (row 1) and getting the key (row 2). The two strategic states for the unlocked door environment correspond to the agent looking for the door (row 1) and making it through the door (row 2). An additional scenario can be found in Appendix G.

For intuition on how a human would use these explanations, consider the cluster in row 1 for the Locked Door. Comparing the first three states in the cluster to the strategic state, a human sees that the policy is suggesting to face the key and move closer to it. As this is a local explanation, it is limited by the initial state being explained as to how close one get to the key. The cluster in row 1 for the Unlocked Door shows that the policy at these states is to face the door. Facing the door within a certain distance seems how the policy breaks down the ultimate strategy. While one might wonder why the strategy is not to get closer to the door (e.g., move up from the second column), recall that the strategic state is explaining the policy and not human intuition. \eat{Lastly, note that for the Unlocked Door, the third state is the same in rows 2 and 3. The rows correspond to explanations for two different initial states, but it is very possible that the same state is encountered in trajectories from each initial state and thus appears in multiple explanations.}\eat{ Such occurrences further illustrate that SSX explanations are local to an initial state.}\\

\begin{figure*}[htbp]
\begin{center}
  \begin{tabular}{ccccc|ccccc}
    \multicolumn{5}{c|}{EAT Scenario 1} &\multicolumn{5}{c}{HUNT Scenario 1}\\ 
   \includegraphics[width=.06\textwidth]{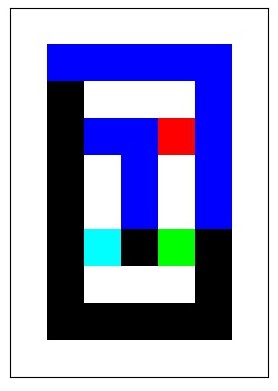}&
   \includegraphics[width=.06\textwidth]{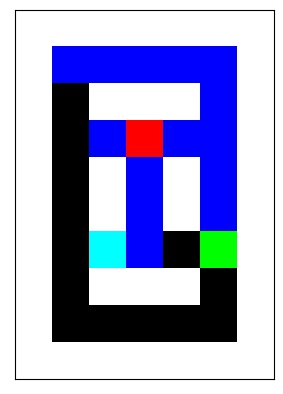}&
   \includegraphics[width=.06\textwidth]{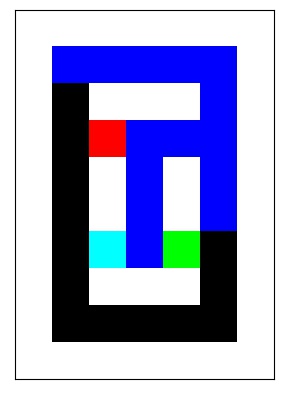}&
   \includegraphics[width=.06\textwidth]{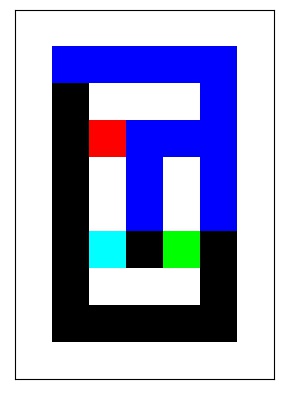}&
   \includegraphics[width=.06\textwidth]{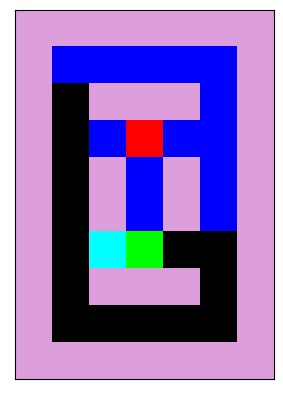}&
   
   \includegraphics[width=.06\textwidth]{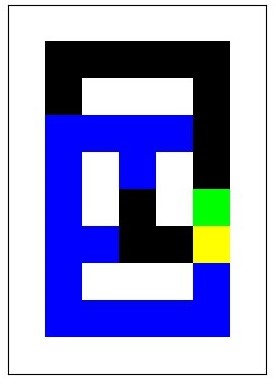} &
   \includegraphics[width=.06\textwidth]{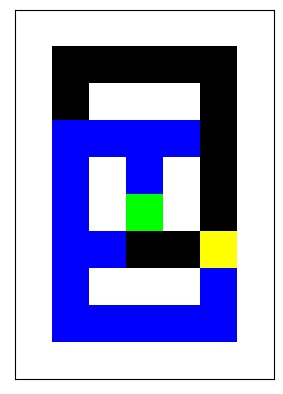} &
   \includegraphics[width=.06\textwidth]{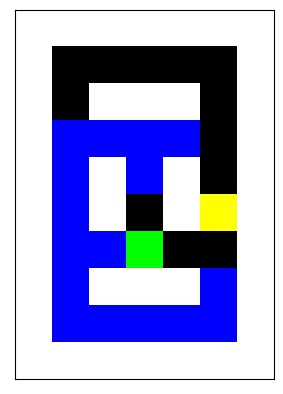} &
   \includegraphics[width=.06\textwidth]{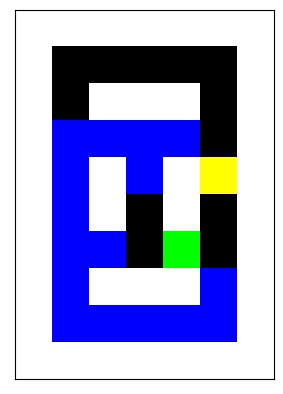} &
   \includegraphics[width=.06\textwidth]{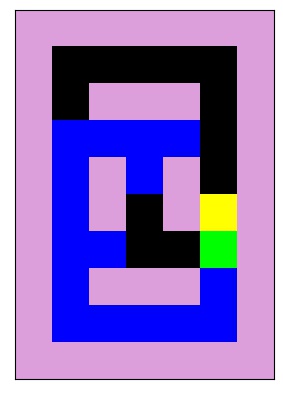}
   \\
   \includegraphics[width=.06\textwidth]{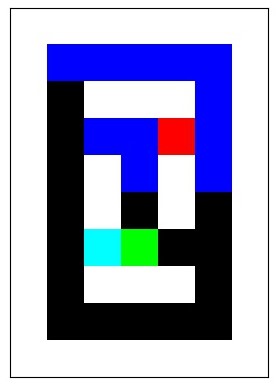} &
   \includegraphics[width=.06\textwidth]{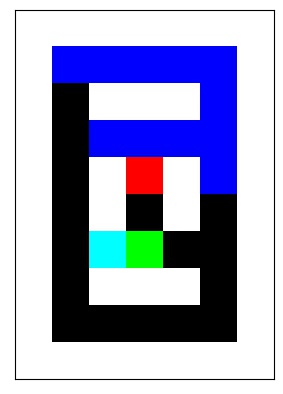} &
   \includegraphics[width=.06\textwidth]{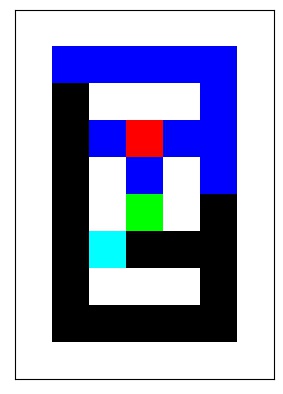} &
   \includegraphics[width=.06\textwidth]{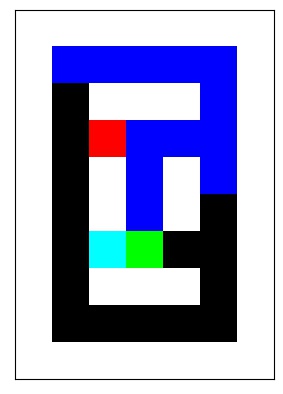} &
   \includegraphics[width=.06\textwidth]{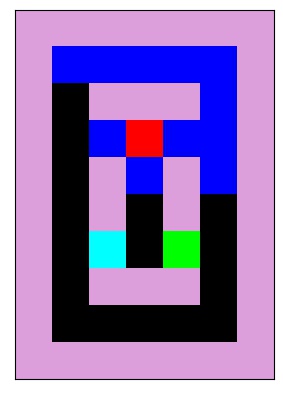} &

   \includegraphics[width=.06\textwidth]{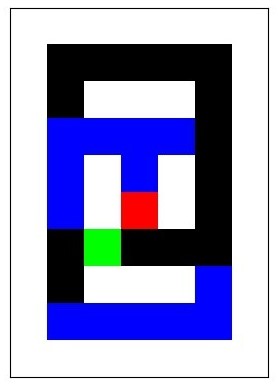} &
   \includegraphics[width=.06\textwidth]{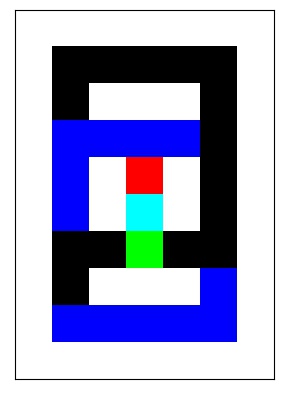} &
   \includegraphics[width=.06\textwidth]{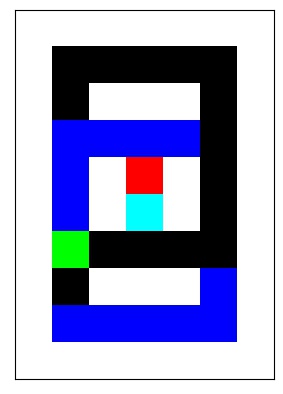} &
   \includegraphics[width=.06\textwidth]{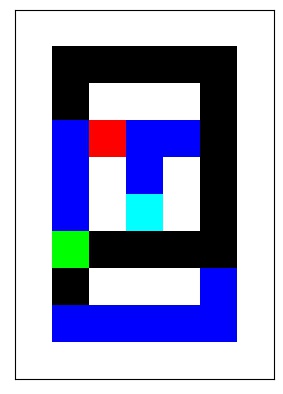} &
    \includegraphics[width=.06\textwidth]{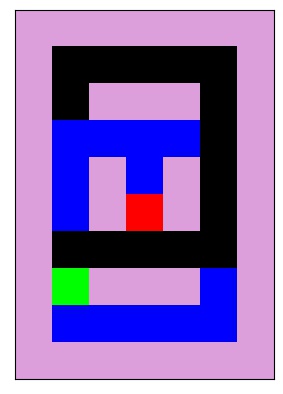}           
\eat{
&&&&&
    \includegraphics[width=.06\textwidth]{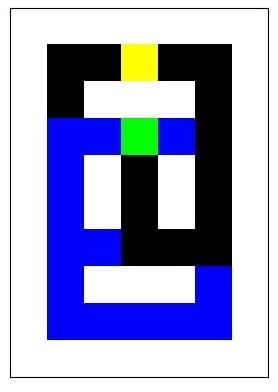} &
   \includegraphics[width=.06\textwidth]{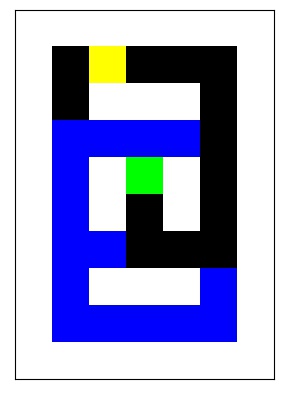} &
   \includegraphics[width=.06\textwidth]{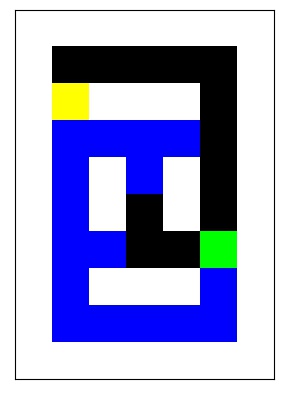} &
   \includegraphics[width=.06\textwidth]{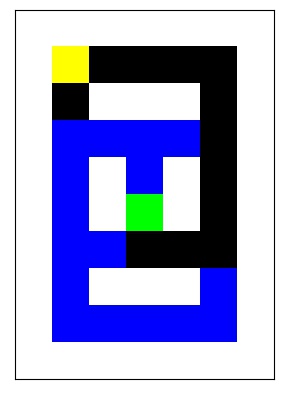} &
    \includegraphics[width=.06\textwidth]{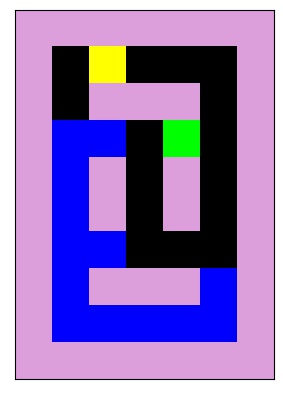} 
}    
\end{tabular}
  \caption{Illustration of our SSX method on minipacman. Two policies, EAT and HUNT, are displayed. Two clusters, one per row, are shown as part of the SSX result. The last board with pink background is a strategic state for each cluster. The color scheme is as follows: green = pacman, red = ghost, yellow = edible ghost, cyan = pill, blue = food, black = food eaten, white/pink=wall. \eat{In EAT scenarios, pacman generally ignores the pill and stays away from the ghost (even if the pill has been eaten). In HUNT, pacman generally looks for the pill (but stays away if the ghost is near it) and moves toward the ghost (if the pill has been eaten).}}
    \label{fig:minipacman_strategic_states}
\end{center}
\vspace{-0.5cm}
\end{figure*}
\noindent\textbf{Minipacman:}
\label{ss-minipacman}
Minipacman is a small version of the classic Pacman game. This game differs from Door-Key with the addition of an adversary - the ghost. The state space is again exponential in the size of the board and the policy is learned as a convolutional neural network with two convolutional and two linear layers\eat{ on a modified environment based on \url{https://github.com/higgsfield/Imagination-Augmented-Agents}}. Two policies are trained with different scenarios. The first scenario, denoted EAT, is for minipacman to eat all the food with no reward for eating the ghost. The second scenario, denoted HUNT, is for minipacman to hunt the ghost with no reward for eating food. 

SSX is again run with local approximations to the state space with the maximum number of steps set to 8. The state space is a $10\times 7$ grid reflecting where food, pacman, a ghost, and the pill are located. Figure \ref{fig:minipacman_strategic_states} displays one sample scenario under both the EAT and HUNT policies, with two meta-states and corresponding strategic states highlighted in pink. In order to interpret the figures, one needs to consider black vs blue pixels. The two strategic states of EAT Scenario 1 show pacman eating the food (row 1), i.e.  columns 2/3 show blue pixels to the right of the pill meaning those pixels were not yet eaten before the strategic state is reached, but then avoiding the ghost and ignoring the pill (row 2). In  HUNT Scenario 1, pacman is either directly moving towards the ghost after having eaten the pill (row 1) or heading away from the pill while the ghost is near it (row 2), i.e. going back to pixels already visited when waiting out the ghost near the pill. Additional scenarios\eat{ for EAT and HUNT} and an experiment with a baseline motivated by \cite{highlights} appear in Appendix H and D, respectively.

\begin{figure}[t]
\begin{center}
   \includegraphics[width=.48\textwidth]{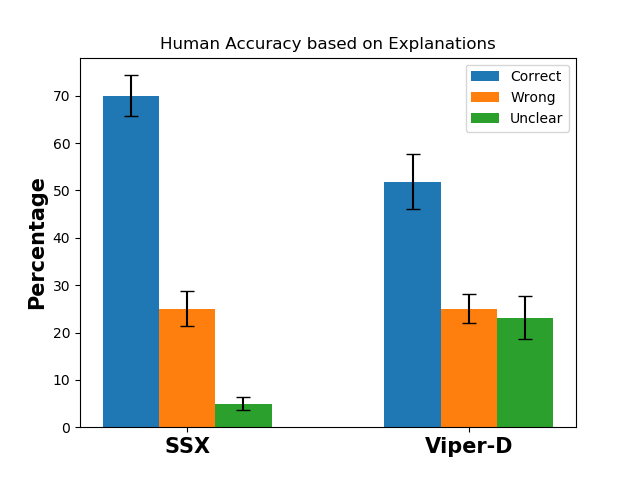}
\end{center}
\vspace{-0.75cm}
  \caption{Above we see the percentage (human) accuracy in predicting if the expert policy is Eat or Hunt based on SSX and Viper-D. Performance difference is statistically significant (paired t-test p-value=0.01). 
  Error bars are 1 std error.}
    \label{fig:userstudy}
    \vspace{-0.5cm}
\end{figure}
\section{User Study}
\label{sec:user_study}

We designed a user study to evaluate the utility of our approach relative to the more standard approach of explaining based on grouping actions. While SSX has thus far been used to give users local explanations about particular scenarios, we use it here to gain insight as to the general goal of a policy because the relevant explanations to compare with are global; as previously discussed, other local literature is about learning inherently explainable models rather than explaining a fixed model or learning contrastive explanations which should be used complementary to our methods. The global applicability of SSX can also be seen as another advantage. As with Four Rooms, we again compare with VIPER -- a state-of-the-art explanation method for reinforcement learning policies -- but use a visual output tailored for the discrete state space and label it Viper-D. We do not compare with methods that output trajectories \cite{highlights} as they require estimating Q-values to determine state importance; while this measure can successfully be used to select important trajectories that give users an idea of what a policy is doing, such important states are not necessarily good representatives of states that one should aim for, as is the goal of strategic states in SSX (see Appendix D for further discussion and related experiments). Among explanation methods, VIPER makes for the best comparison as it requires a similar amount of human analysis of the explanation (by observing states), and while meant for global explainability, also gives local intuitions, as opposed to other global methods. The utility of each approach is measured through a task posed to study participants: users must guess the intent of the expert policy based on provided explanations which are either output by SSX or VIPER. Such a task oriented setup for evaluation is heavily encouraged in seminal works on XAI \cite{rsi,lipton2016mythos,tip}. 

\noindent\textbf{Setup:} We use the minipacman framework with the EAT and HUNT policies trained above and each question shows either an SSX explanation or Viper-D explanation and asks the user ``Which method is the explanation of type A (or B) explaining?" to which they must select from the choices Hunt, Eat, or Unclear. Methods are anonymized (as A or B) and questions for each explanation type are randomized. Ten questions (five from both the EAT and HUNT policies) are asked for each explanation type giving a total of twenty questions to each participant. At the end of the study, we ask users to rate each explanation type based on a 5-point Likert scale for four qualitative metrics - completeness, sufficiency, satisfaction and understandability - as has been done in previous studies on explainable RL \cite{actinfl}. For users to familiarize themselves with the two types of explanations we also provided training examples at the start of the survey, one for each type. 

To be fair to VIPER explanations, rather than just displaying rules in text which may not be aesthetically pleasing, we created a visualization which not only displayed the (five) rules to the user, but also three boards, one each for pacman, the ghost, and the pill, highlighting their possible locations as output by the rule. This visualization, which we call Viper-D, is beyond the typical decision tree offered by VIPER and better renders explanations in our discrete setting. 
Screenshots of sample visualizations along with the instruction page and optional user feedback can be found in Appendix I.

The study was implemented using Google Forms and we received 37 responses from people with quantitative/technical backgrounds, but not necessarily AI experts. We removed 5 responses as they were likely due to users pressing the submit button multiple times as we twice received multiple answers within 30 seconds that were identical. 

\noindent\textbf{Observations:} Figure \ref{fig:userstudy} displays user accuracy on the task for method SSX and Viper-D. Users were able to better distinguish between the EAT and HUNT policies given explanations from SSX rather than Viper-D and the difference in percentage correct is statistically significant (paired t-test p-value is 0.01). Another interesting note is that less than 5\% of SSX explanations were found to be Unclear whereas more than 25\% of Viper-D explanations were labeled Unclear, meaning that, right or wrong, users felt more comfortable that they could extract information from SSX explanations. See Appendix I for results of qualitative questions to which users scored SSX higher than VIPER.

\vspace{-0.2cm}
\section{Discussion}
\label{sec:disc}
\vspace{-0.15cm}
We have seen in this work that our novel approach of identifying strategic states leads to more complete, satisfying and understandable explanations, while also conveying enough information needed to perform well on a task. Moreover, it applies to single agent as well as multi-agent adversarial games with large state spaces. Further insight could be distilled from our strategic states by taking the difference between the variables in some particular state and the corresponding strategic state and conveying cumulative actions an agent should take to reach those strategic states (viz. go 2 steps up and 3 steps right to reach a door in Four Rooms). This would cover some information conveyed by typical action-based explanations\eat{ we have seen} while possibly enjoying benefits of both perspectives. Other future directions include seeing if strategic states could be used as intermediate goals for efficiently training new policies and extending our idea to continuous state spaces.\eat{ While one could discretize the state space which could be suboptimal, it would be interesting to see if it can be symbiotically done.}

\bibliography{ExAbsent}
\bibliographystyle{plain}
\appendix

\section*{Appendix}
\setcounter{section}{0}

\section{Algorithmic Details}
We first prove Proposition 1.
\begin{proof}[Proof]
Consider two sets $U$ and $V$ consisting of strategic states of meta-state $\Phi$, where $U\subseteq V$. Let $w$ be a strategic state $\not\in V$ and $G_{\Phi}(.)$ represent the objective in equation \ref{eq:strategic}, then we have
\begin{align}
\label{eq:U}
    G_{\Phi}(U\cup w)-G_{\Phi}(U)&\\
    = C(w, \Phi)&-\lambda\sum_{u\in U}\max\left( \gamma(w,u),\gamma(u,w)\right)\nonumber
\end{align}
Similarly,
\begin{align}
\label{eq:V}
    G_{\Phi}(V\cup w)-G_{\Phi}(V) &\\
    = C(w, \Phi)
    &-\lambda\sum_{v\in V}\max\left( \gamma(v,w),\gamma(w,v)\right)\nonumber
\end{align}
Subtracting equation \eqref{eq:V} from \eqref{eq:U} we get,
\begin{align}
    \eqref{eq:U}-\eqref{eq:V}&=\lambda\left(\sum_{v\in V}\max\left( \gamma(v,w),\gamma(w,v)\right)\right.\\
    &\quad\quad\quad\quad\left.-\sum_{u\in U}\max\left( \gamma(w,u),\gamma(u,w)\right)\right)\nonumber\\
    &= \lambda\sum_{v\in V\setminus U}\max\left( \gamma(v,w),\gamma(w,v)\right)\ge 0\nonumber
\end{align}
Thus, the function $G_{\Phi}(.)$ has diminishing returns property.
\end{proof}

We next comment on the convergence for Algorithm \ref{algo:meta-state} which follows directly since our objective in equation (\ref{eq:mainobj}) is bounded and monotonically decreases at each iteration.
\begin{proposition}
Meta-state finding Algorithm \ref{algo:meta-state} converges.
\end{proposition}
\eat{
\begin{proof}[Proof Sketch]
Our objective is bounded and in every iteration of our algorithm the objective value monotonically decreases. These two properties imply that our algorithm will converge.
\end{proof}
}

\begin{figure}[h]
\centering
\begin{tabular}{c}
   \includegraphics[width=.45\textwidth]{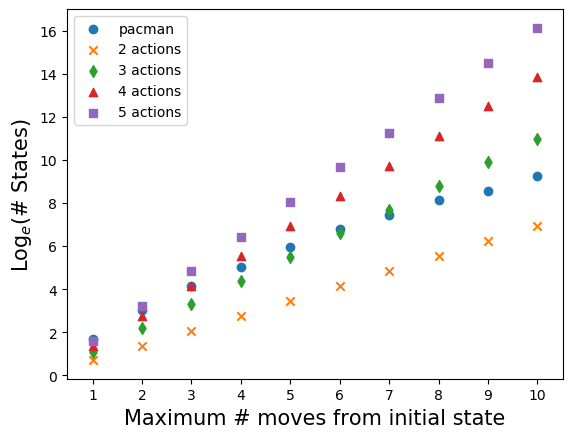}
\end{tabular}
  \caption{Illustrations of SSX state space size in minipacman. Worst case state space size for local approximations is $N^M$ where $N$ is the maximum number of moves made and $M$ is the number of possible actions per move. Pacman's state space is averaged over 100 random samples for each $N=1,\ldots,10$. The state space of minipacman, while also growing exponentially, grows much slower (like a game with 2-3 actions per move) which makes SSX a practical method for such games.}
    \label{fig:state_space_size}
    \vspace{-0.25cm}
\end{figure}

\section{Additional practical considerations}
\noindent\textbf{Additional information on scalability:}
Here we discuss additional details to those on scalability discussed in Section \ref{ss-practical_considerations} and show Figure \ref{fig:state_space_size} which displays how the state space size in minipacman, discussed in Section \ref{ss-minipacman}, grows in practice as the number of possible moves $N$ allowed for the local approximation grows. At any position on the board, minipacman has at most 4 possible actions (3 possible directions to move or stay) and the ghost has an additional 3 potential actions for a total of 7 possible state movements at most. The state space of minipacman is averaged over 100 random samples for each $N=1,\ldots,10$. Also note that the size of the local approximation to the state space will not be affected if the board size increases because only local states are considered.

\begin{figure*}[h]
\centering
\begin{tabular}{cc}
   \includegraphics[width=.45\textwidth]{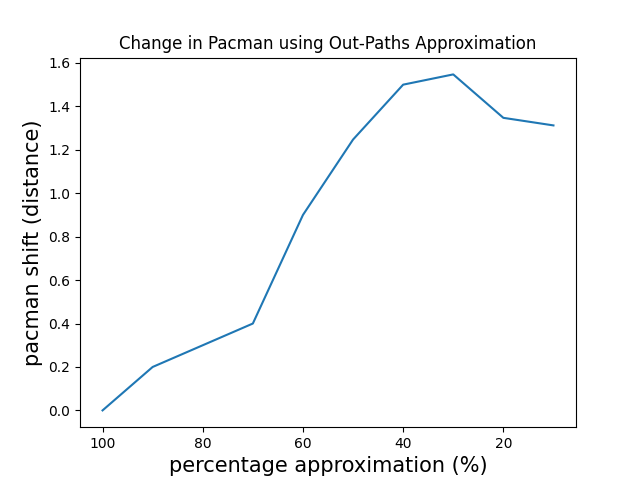} &    \includegraphics[width=.45\textwidth]{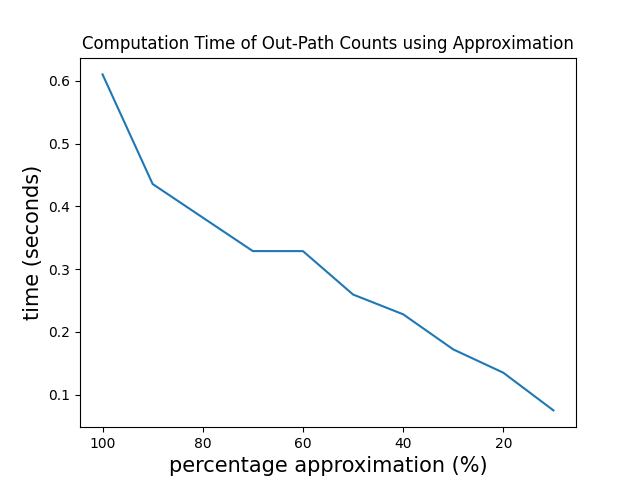}
\end{tabular}
  \caption{The left figure shows the (average) change in pacman position from the original as the percentage of states used in approximating the count of out-paths ($C(s, \Phi)$ decreases. The right figure correspondingly shows the time for a single computation of $C(s, \Phi)$ as the percentage of states used in the approximation decreases. A change of distance in pacman $< 1$ indicates an average movement for pacman of less than one space on the grid. We observe a reduction of approximately 50\% in computation time of $C(s, \Phi)$ maintains stability in that the pacman position in the primary strategic state has moved less than one space on average.
}
    \label{fig:out_paths_approximation}
\end{figure*}

\noindent\textbf{Approximating Counts of Out-Paths:}
Computing the number of out-paths, $C(s, \Phi)$ (defined in equation \ref{eq:pathcount}), is perhaps the most expensive and critical aspect of our algorithm as it is used both as a regularization when learning the meta-states in equation (\ref{eq:mainobj}) as well as a criterion to identify strategic states in equation (\ref{eq:strategic}). We have already noted that approximating it (through sampling) might lead to significant computational savings. However, the main question is how sensitive is our approach to such approximations? We study this next. 
As seen in equation (\ref{eq:pathcount}), the computation of out-paths involves searching over all paths between states in each meta-state with those states in other meta-states. We consider a speedup where we randomly sample only a fixed fraction of the states in other meta-states. Figure \ref{fig:out_paths_approximation}
displays results illustrating that approximating the count of out-paths can maintain stability while offering significant computational savings. Specifically, we observe a reduction of approximately 50\% in computation time of $C(s, \Phi)$ maintains stability in that the pacman position in the primary strategic state has moved less than one space on average.

\noindent\textbf{Storing Paths:}
The predecessor matrix $P$ is defined such that $P_{ij}$ is the predecessor of state $j$ on a shortest path from $i$ to $j$ (and infinity if no such path exists). 
\eat{\[
P_{ij}=\begin{cases}
\mbox{NIL if $i=j$ or no path exists from $i$ to $j$}\\
\mbox{predecessor of $j$ on some shortest path from $i$ to $j$} 
\end{cases}
\]}
This matrix is used to retrieve the shortest path between any two states $i$ and $j$. Then a strategic state is defined as a state $s'$ such that $P_{st}=s'$ where $\phi(s)=\phi(s')\ne\phi(t)$, i.e. $s'$ is the last node on the shortest path between states $s$ and $t$ that are in two different meta-states that lies in the same meta-state as $s$.  Then, by this definition, we can penalize the number of strategic states.

\noindent\textbf{Tractability of $\Gamma$:} $\Gamma$, along with a predecessor matrix $P$ that can be used to derive the shortest paths, can be computed using Dijkstra's shortest path algorithm in $O(|\mathcal{S}|^2\log{|\mathcal{S}|)}$ because all edge weights are non-negative. Note that computation of $\Gamma$ means that SSX requires access to a policy simulator for $\pi_E$, and in practice, might require simulation for estimation when $\Gamma$ cannot be computed exactly. This is a common requirement among related explanation methods, e.g., in order to simulate important trajectories \cite{highlights} or samples to train a decision tree \cite{viper}. Computation of the entire $\Gamma$ matrix may be computationally costly and unnecessary in certain environments, particularly when an expert policy only dictates an agent to visit a small subset of the possible states. In such cases, $\Gamma$ can be restricted only to regions of the state space that are more likely, resulting in improved efficiency. \eat{Note that the predecessor matrix was already computed for matrix $\Gamma$ above. }One may also consider the likelihood (rather than count) of out-paths by replacing the indicator in eq. (\ref{eq:pathcount}) with $\gamma(s',s'')$.

\noindent\textbf{Eigen Decompositions:} We use the python function scipy.sparse.linalg.eigsh which is a wrapper to ARPACK \cite{arpack} functions that rely on Fortran77 subroutines and the use of these ARPACK functions is the classic method for large-scale eigenvalue problems. Note two important aspects about the computation: 1) SSX only needs $k$ eigenvectors where $k$ is the number of clusters desired which is typically small, and 2) the Laplacian of is typically highly sparse which makes it amenable to algorithms tailored for sparse matrices like the one we use. For sparse matrices, the algorithm upon which ARPACK is based has complexity $O(nk^2)$ where $n$ is the matrix size and $k$ is the number of eigenvectors desired \cite{stathopoulos97}.

\noindent\textbf{Number of Meta-states $k$:} The number of meta-states can be chosen using standard techniques as trying different $k$ and finding the knee of the objective (i.e. where the objective has little improvement) or based on domain knowledge. State representations may affect the (appropriate) number.

\begin{figure*}[t]
\begin{center}
  \begin{tabular}{cccccc}
    \multicolumn{6}{c}{Expert model trained on 50000 episodes}\\ 
   \includegraphics[width=.1\textwidth]{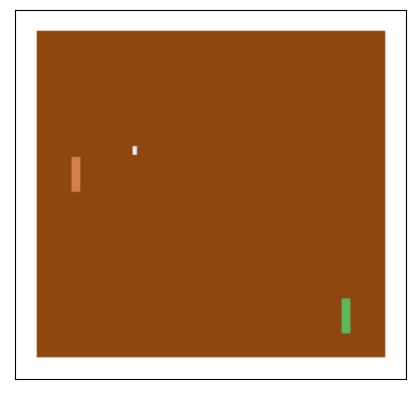} &
    \includegraphics[width=.1\textwidth]{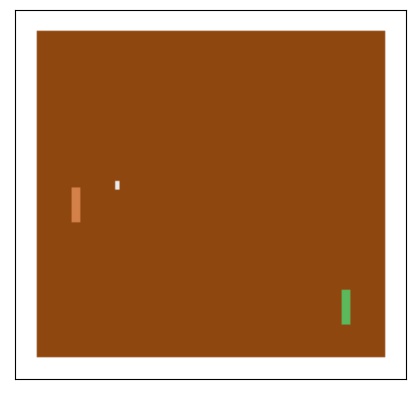} &\includegraphics[width=.1\textwidth]{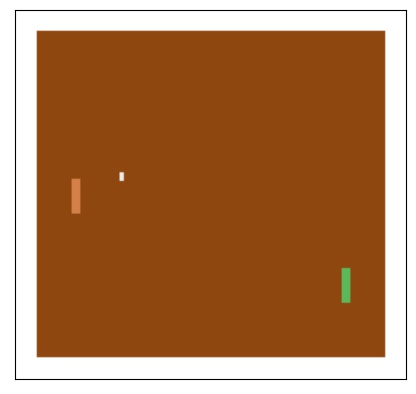} &\includegraphics[width=.1\textwidth]{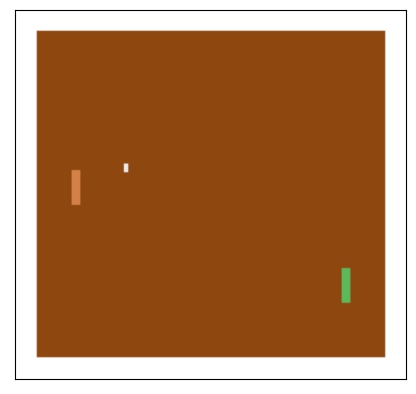} &\includegraphics[width=.1\textwidth]{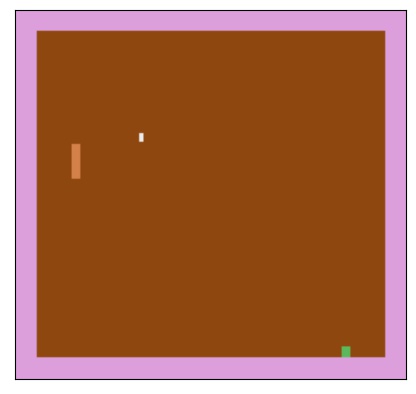}  &
    \\
   \includegraphics[width=.1\textwidth]{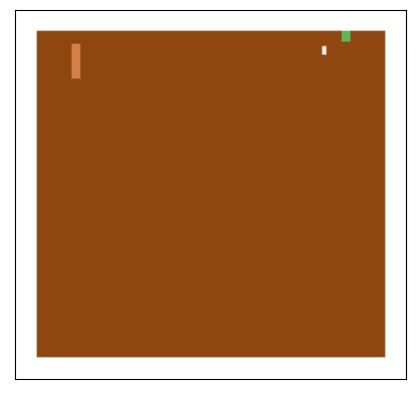} &
    \includegraphics[width=.1\textwidth]{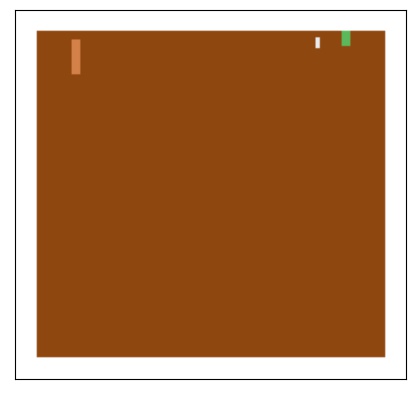} &
    \includegraphics[width=.1\textwidth]{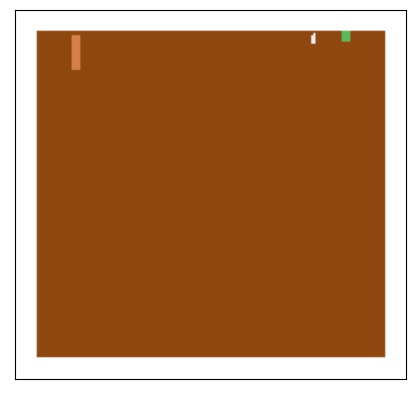} &
    \includegraphics[width=.1\textwidth]{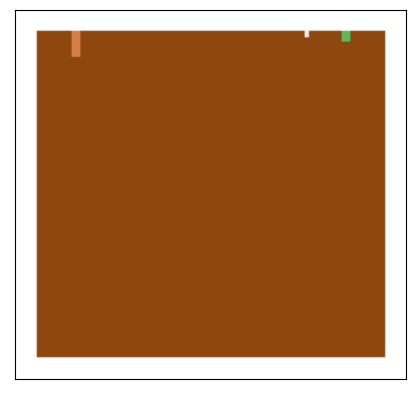} &
    \includegraphics[width=.1\textwidth]{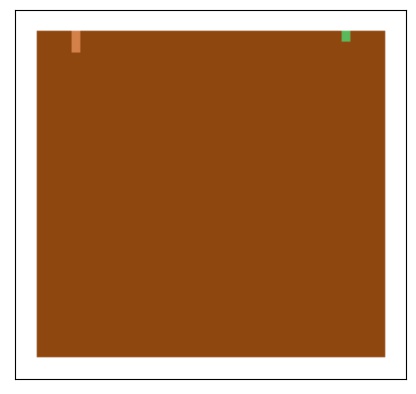} &
    \includegraphics[width=.1\textwidth]{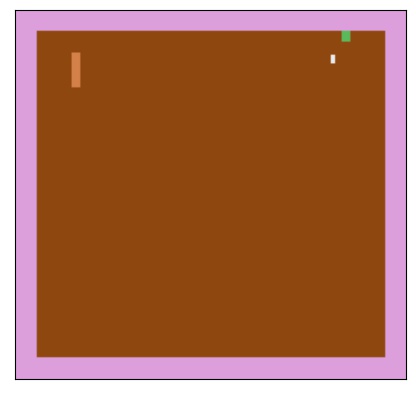} 
    \end{tabular}
  \caption{Illustration of selecting strategic states on pong. Each row is a cluster from a different scenario and last board with pink background is a strategic state for that cluster. Green bar is the agent and organge bar is the adversary.}
    \label{fig:pong_strategic_states}
\end{center}
\end{figure*}

\section{Pong Example}
Pong is another adversarial game that we apply SSX to. In this game, one controls a paddle and must return the ball to the other side of the game with the goal of hitting the ball past the adversary's paddle on the other side of the board. The ball can bounce of the walls at the top and bottom of the board. This example differs from the minipacman example in that the user has no access to the adversary, meaning that computing the maximum likelihood path matrix $\Gamma$ requires simulation. We make use of the OpenAI gym atari implementation \cite{gym} and apply SSX to a policy learned from a neural network with one hidden layer with 200 neurons trained with a stochastic policy gradient algorithm (\url{https://github.com/omkarv/pong-from-pixels}) with 50,000 episodes so that the agent outperforms the adversary. The state space is an $80\times 80$ grid and input to the policy is the difference between the current and previous states, which allows for a notion of motion (e.g., knowledge of which direction the ball is heading). Local approximations to the input state space are used with the maximum number of steps set to 8.

Results are illustrated in Figure \ref{fig:pong_strategic_states}. The two rows in the figure are from applying SSX to two different scenarios, where the green bar is the agent and the orange bar is the adversary. In the first (row) scenario, the ball is traveling to the upper wall away from the opponent, who is in the upper half of the board. The explanation suggests that the policy dictates the agent to stay close to the bottom of the board. This makes sense because from this position, the agent can wait to see where the adversary moves before deciding at what angle to hit the ball. In the second (row) scenario, the ball is also moving to the wall, but closer to the agent, and both the agent and adversary are in the upper part of the board. In this scenario, the strategy is to remain above the ball in order to hit the ball at an angle pushing the ball to the bottom of the board (away from the adversary).

\begin{figure*}[t]
\begin{center}
  \begin{tabular}{ccccc|ccccc}
    \multicolumn{5}{c|}{EAT} &\multicolumn{5}{c}{HUNT}\\ 
   \includegraphics[width=.06\textwidth]{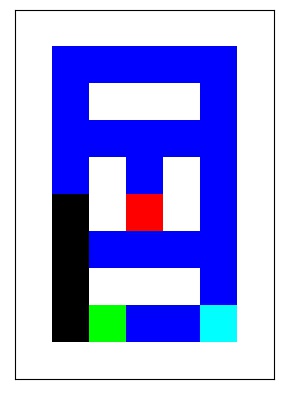} &
   \includegraphics[width=.06\textwidth]{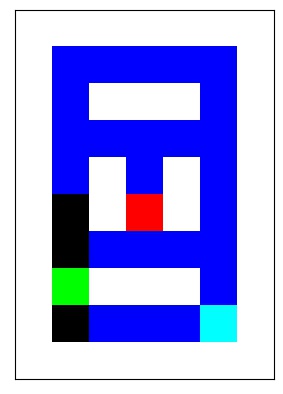} &
   \includegraphics[width=.06\textwidth]{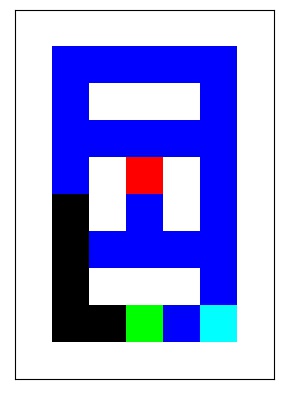} &
   \includegraphics[width=.06\textwidth]{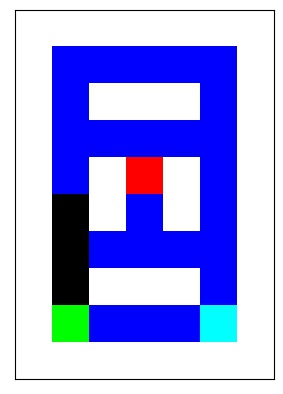} &
   \includegraphics[width=.06\textwidth]{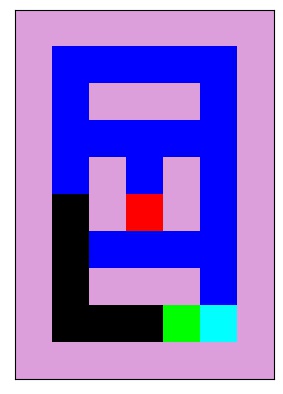} &
   
   \includegraphics[width=.06\textwidth]{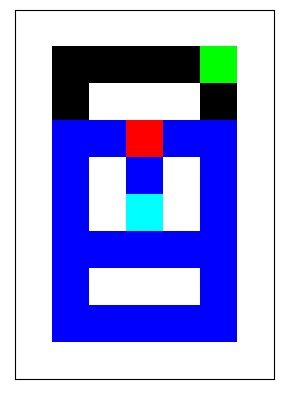} &
   \includegraphics[width=.06\textwidth]{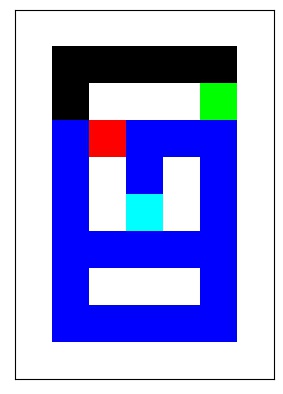} &
   \includegraphics[width=.06\textwidth]{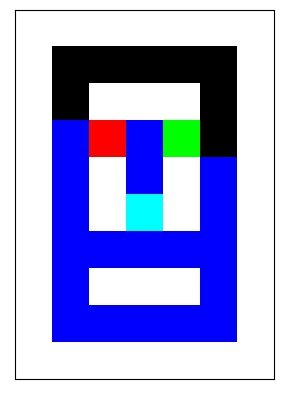} &
   \includegraphics[width=.06\textwidth]{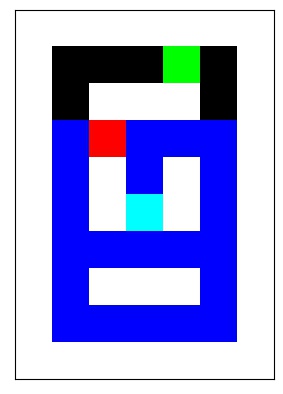} &
   \includegraphics[width=.06\textwidth]{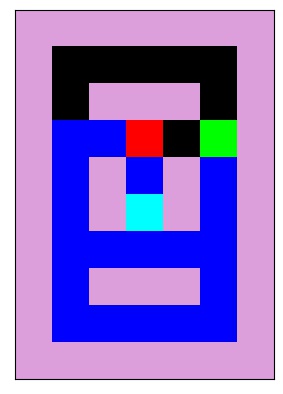} 
   \\
   \includegraphics[width=.06\textwidth]{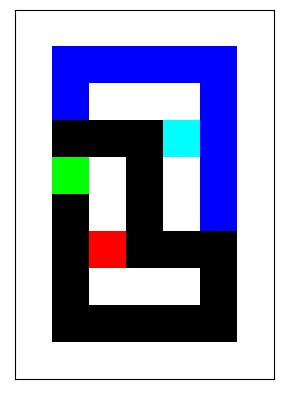} &
\includegraphics[width=.06\textwidth]{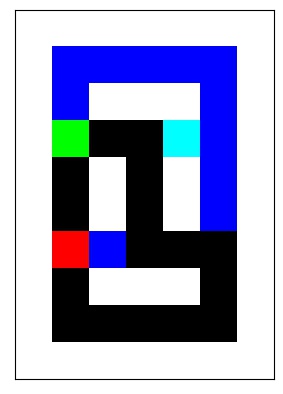} &
\includegraphics[width=.06\textwidth]{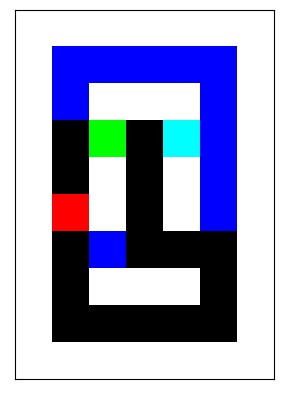} &
\includegraphics[width=.06\textwidth]{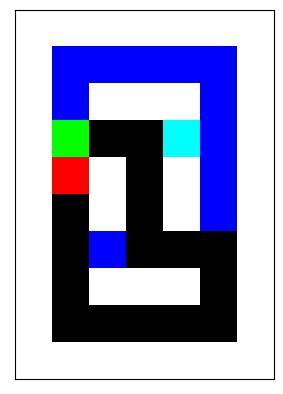} &
\includegraphics[width=.06\textwidth]{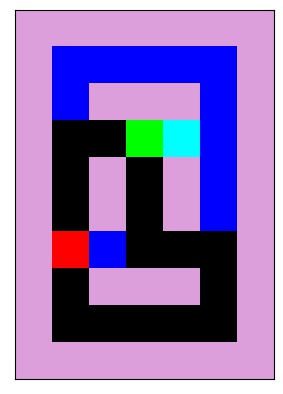} &
   
   \includegraphics[width=.06\textwidth]{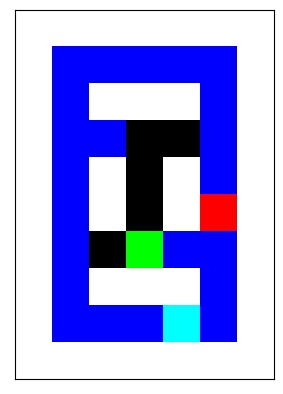} &
   \includegraphics[width=.06\textwidth]{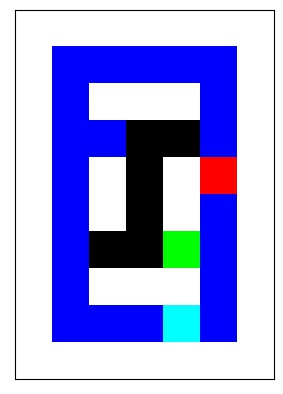} &
   \includegraphics[width=.06\textwidth]{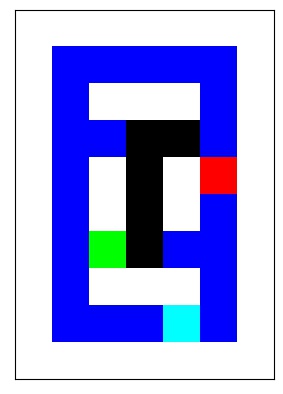} &
   \includegraphics[width=.06\textwidth]{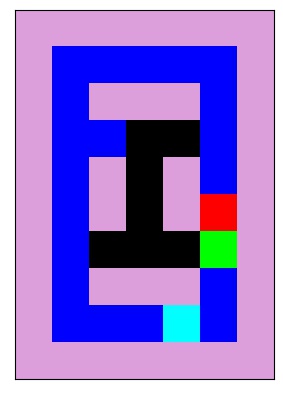} &
    \end{tabular}
  \caption{Illustration of selecting important states on minipacman. Two policies, EAT and HUNT, are displayed across two scenarios for each. For each scenario, a single cluster is shown. For a given cluster, the last board with pink background is an important state for that cluster as defined by equation (\ref{eq:importance_value}). The color scheme is as follows: green = pacman, red = ghost, yellow = edible ghost, cyan = pill, blue = food, black = food eaten, white/pink=wall.}
    \label{fig:minipacman_important_states}
\end{center}
\end{figure*}

\section{Comments on State Importance}
A potential baseline for explanations is motivated by the policy summarization method \cite{highlights}. Explanations in that work are offered as simulated trajectories that are deemed important. Importance of states along the trajectory is defined by 
\[
\label{eq:importance_value}
I(s) = \max_a{Q^{\pi_E}_{(s,a)}} - \min_a{Q^{\pi_E}_{(s,a)}},
\]
where $Q^{\pi_E}_{(s,a)}$ is the Q-value for policy $\pi_E$ for taking action $a$ at state $s$. By definition, this gives a measure of variation (by action) for a given state, i.e., the large the potential impact of which action is taken, the higher the importance value. While this measure can successfully be used to select important trajectories that give users an idea of what a policy is doing, as done in \cite{highlights}, such important states are not necessarily good representatives of states that one should aim for, as is the goal of strategic states in SSX. Figure \ref{fig:minipacman_important_states} depicts a few examples to demonstrate this property on minipacman. Note that there is a negative reward for eating a pill in the EAT scenario because the policy is to eat all the regular food while avoiding the ghost. Q-values were estimated using the equation, Q(s,a) = r(s,a) + E[V(t)], where $r(s,a)$ is the reward for taking action $a$ in state $s$ and is determined by the scenario (EAT or HUNT), $t$ is the state that pacman transitions to by taking taking action $a$ in state $s$, and $V(t)$ is the value of being in state $t$. The expectation is taken with respect to the new state (because the ghost moves stochastically after pacman takes action $a$).

In both EAT scenarios, important states are found with pacman located directly next to the food; indeed, these are important according to the measure in equation (\ref{eq:importance_value}) because the difference of whether pacman moves away from the pill or eats it is large. However, these are not positions where one would expect to guide pacman in the EAT scenario. Similarly, for the HUNT scenarios, the important states are found where pacman is relatively close to the ghost. For example, in the first row, if pacman moves left, there is a chance the ghost will move right and eat pacman, whereas if pacman moves up or down, pacman will continue. 

\begin{table}[htbp]
\centering
\begin{tabular}{|c|c|c|c|c|}
\multicolumn{5}{c}{Minipacman} \\ \hline
$N$ &3 &4 &5 &6  \\ \hline
3&0 &0.70  &0.92  &1.14   \\ \hline
4&- &0  &0.85  &0.78   \\ \hline
5&- &-  &0  & 1.29  \\ \hline
\end{tabular}
\\
\begin{tabular}{|c|c|c|c|c|}
\multicolumn{5}{c}{Ghost} \\ \hline
$N$ &3 &4 &5 &6  \\ \hline
3&0 &0.88  &1.05  &2.59   \\ \hline
4&- &0  &1.36  &2.23   \\ \hline
5&- &-  &0  &1.73   \\ \hline
\end{tabular}
\\
\begin{tabular}{|c|c|c|c|c|}
\multicolumn{5}{c}{Food} \\ \hline
$N$ &3 &4 &5 &6  \\ \hline
3&0 & 0.52 &0.73  &0.84   \\ \hline
4&- &0  &0.57  &0.93   \\ \hline
5&- &-  &0  &0.88   \\ \hline
\end{tabular}
\caption{Faithfulness measures for size of local neighborhood. For Minipacman, measures are distances of minipacman positions in strategic states when using different local approximations with varying approximation  size $N$, where $N$ is the maximum number of steps allowed for a state to be included. Distances are symmetric (hence use of -).}
\label{tab:faithfulness_neighborhood}
\end{table}

\section{Faithfulness and Stability of Local Approximations}
We here investigate how sensitive strategic states in SSX are to the size of the local approximation as well as minor changes in the initial state. Let $N$ be the maximum number of steps that can be taken (as referred to above in the discussion on Scalability in Section 3.4). We allow $N$ to vary from 3 to 6 steps and run SSX for the minipacman setup from various starting boards along a trajectory of the HUNT scenario. For each trial, we take the priority strategic state of the cluster containing the initial board and compare it against trials from the same initial board but with different $N$. We consider three distance metrics for each comparison: the distance between minipacman positions, the distance between ghost positions, and the distance between remaining food indicators of the boards. 

\begin{table}[t]
\center
\begin{tabular}{|c|c|}
\hline
 & Avg Distance \\ \hline
Minipacman & 0.63 \\ \hline
Ghost & 1.61 \\ \hline
Food & 1.32 \\ \hline
\end{tabular}
\caption{Stability measures for perturbations of the initial state. For Minipacman, measures are distances of minipacman positions in strategic states when using different initial states (comparing a strategic state from initial state to that of a perturbed initial state).}
\label{tab:consistency_neighborhood}
\end{table}

Table \ref{tab:faithfulness_neighborhood} displays the results. Distance is measured using $l_2$ norm on the difference of (x,y) coordinates for minipacman and the ghost and the difference of indicators of whether or not food is present for the food. For reference, a value of 1 for minipacman means that minipacman was usually 1 position away (horizontally or vertically) in the two strategic states being compared. Values close to 0 mean that minipacman was often in the same spot in respective strategic states. We first note that distance generally increases the larger the difference of $N$ between trials, as expected. The second note is that the distance of the ghost is larger than that of minipacman, which is also expected as the ghost moves randomly whereas minipacman is driven by the policy.

We next consider the stability of local approximations by comparing strategic state results from an initial state to that when slightly perturbing the initial state. Stability is illustrated on minipacman, where perturbations are done by randomly removing 3 pieces of food from the initial state being considered by SSX. For each initial state, 10 additional random perturbations are used, and various initial states along a trajectory are used. Distances are measured as described above for minipacman, the ghost, and the food. Results are given in Table \ref{tab:consistency_neighborhood}. Again, minipacman on average is less than one position away when comparing SSX results between two initial states that differ by a minor perturbation. As expected, the ghost moves more since the ghost is not controlled by the policy. The distance of 1.32 for food corresponds to a difference of close to 3 pieces of food (since distance is an $l_2$ norm between the indicator matrices of food). This makes sense as the perturbations make the boards differ by 3 units of food. 

\begin{figure}[t]
\begin{center}
   \includegraphics[width=.48\textwidth]{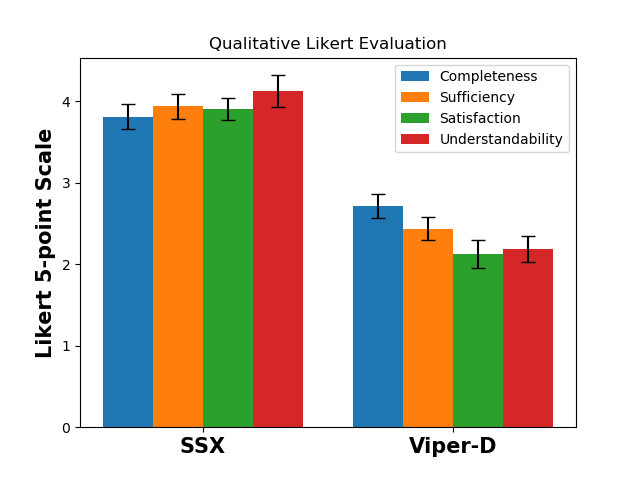}
\end{center}
  \caption{Above we see a 5-point Likert scale (higher better) for four qualitative metrics used in previous studies \cite{actinfl}. The difference is statistically significant for all four metrics (p-values $< 2\times 10^{-5}$ \eat{$1.4\times 10^{-5}$, $1.6\times 10^{-8}$, $5.2\times 10^{-10}$ and $9.5\times 10^{-9}$ respectively}). 
  Error bars are 1 std error.}
    \label{fig:userstudy_likert}
\end{figure} 

\begin{table*}[htbp] 
\centering
\caption{Parameters used for Four Rooms, Door-Key, Minipacman, and Pong experiments}
\label{tab:hyperparameters}
\begin{tabular}{|c|c|c|c|c|}
\hline
& \multicolumn{4}{c|}{\textbf{Domain}} \\ \hline
\textbf{Parameter} & Four Rooms & Door-Key & Minipacman&Pong \\ \hline
\# strategic states $k$ & 2 & 5& 5&5\\ \hline
$\lambda$ from eq. (\ref{eq:strategic})& 50.0 & 1.0 & 0.1&0.1 \\ \hline
$\epsilon_g$ from algorithm \ref{algo:SS}& 0.1 & 0.1 & 0.1&0.1\\ \hline
$N$ \# steps used for & NA & 6 & 6 &8\\ 
local approximation to $\mathcal{S}$ & & & &\\ \hline
\end{tabular}
\end{table*}

\section{Reproducing the Experiments}
Training of all models uses default parameters from the respective github repositories used for each environment (links in Section \ref{sec:exp}). Parameters used for the experiments are given in Table \ref{tab:hyperparameters}.

\section{Additional DoorKey Examples}
\begin{figure*}[t]
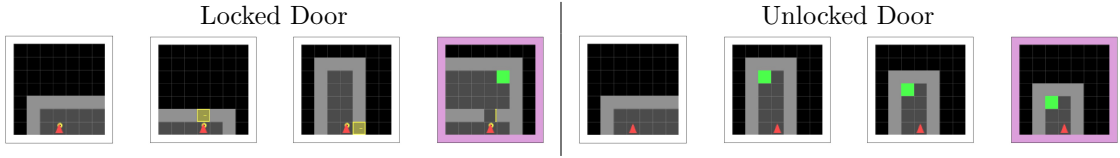

\begin{center}
  \begin{tabular}{cccc|cccc}
    \multicolumn{4}{c|}{Locked Door} &\multicolumn{4}{c}{Unlocked Door}\\ 

\includegraphics[width=.09\textwidth]{figs/DoorKey_21/DoorKey_9_of_21steps_Cluster1_Row0_Column0.jpg} &
   \includegraphics[width=.09\textwidth]{figs/DoorKey_21/DoorKey_9_of_21steps_Cluster1_Row0_Column3.jpg} &
   \includegraphics[width=.09\textwidth]{figs/DoorKey_21/DoorKey_9_of_21steps_Cluster1_Row0_Column4.jpg} &
   \includegraphics[width=.09\textwidth]{figs/DoorKey_21/DoorKey_9_of_21steps_Cluster1_Row0_Column5.jpg} &
   \includegraphics[width=.09\textwidth]{figs/DoorKey_unlocked_21/DoorKey_unlocked_9_of_21steps_Cluster0_Row0_Column0.jpg} &
   \includegraphics[width=.09\textwidth]{figs/DoorKey_unlocked_21/DoorKey_unlocked_9_of_21steps_Cluster0_Row0_Column2.jpg} &
   \includegraphics[width=.09\textwidth]{figs/DoorKey_unlocked_21/DoorKey_unlocked_9_of_21steps_Cluster0_Row0_Column3.jpg} &
   \includegraphics[width=.09\textwidth]{figs/DoorKey_unlocked_21/DoorKey_unlocked_9_of_21steps_Cluster0_Row0_Column4.jpg}
 
   \end{tabular}
     \caption{Illustration of our SSX method on an additional scenario for Door-Key. Policies were trained on two different environments: Locked Door and Unlocked Door.  Each row corresponds to a meta-state and strategic state (outlined in pink) from running SSX starting at a different number of moves into the same path (one path for completing the task in each of the two environments). For the Locked Door environment, the agent uses the key to open the door. For the Unlocked Door environment, the agent proceeds to the goal state (after having already entered the room) in green.}
    \label{fig:doorkey_strategic_states_supp}
  \end{center}
  \vspace{-0.5cm}
 \end{figure*}
Figure \ref{fig:doorkey_strategic_states_supp} shows results of SSX applied to an additional scenarios for the Door-Key example. These follow the corresponding scenarios from the main paper. In the Locked Door environment, the agent uses the key to open the door, having previously found the found. In the Unlocked Door environment, the agent proceeds to the goal state after having previously entered the room.

\section{Additional Minipacman Examples}
\begin{figure*}[h]
\begin{center}
  \begin{tabular}{ccccc|ccccc}
\multicolumn{5}{c|}{EAT Scenario 2} &\multicolumn{5}{c}{HUNT Scenario 2}\\ 

   \includegraphics[width=.06\textwidth]{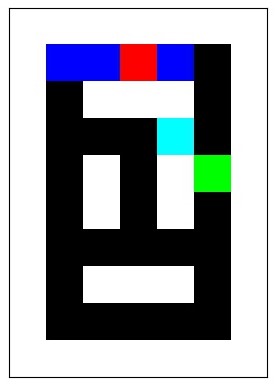} &
    \includegraphics[width=.06\textwidth]{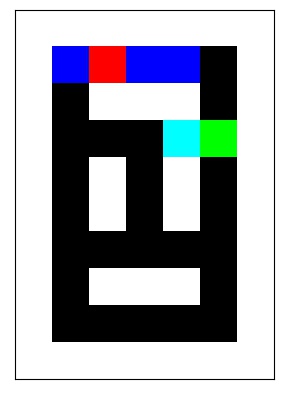} &\includegraphics[width=.06\textwidth]{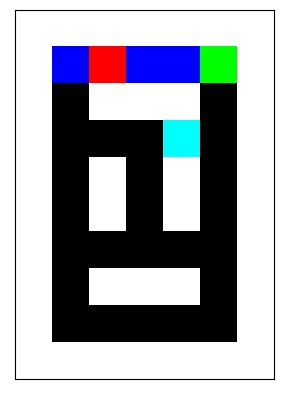} &\includegraphics[width=.06\textwidth]{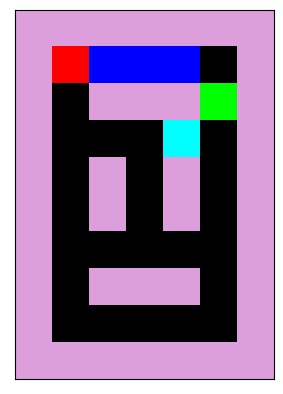} & &
   
   \includegraphics[width=.06\textwidth]{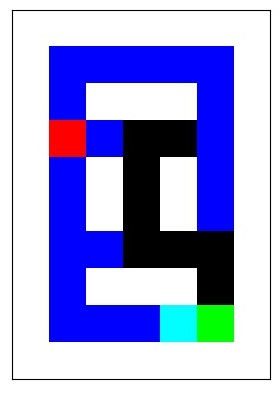} &
    \includegraphics[width=.06\textwidth]{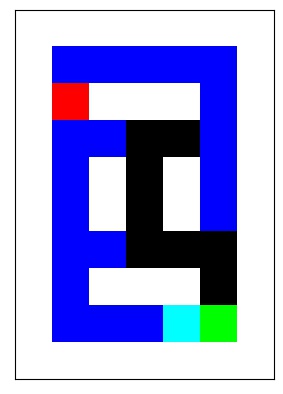} &
    \includegraphics[width=.06\textwidth]{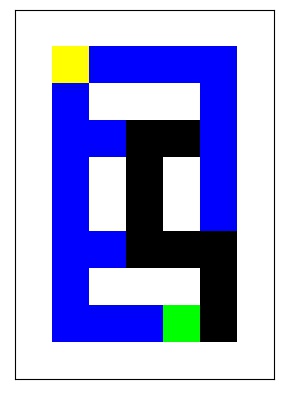} & 
    \includegraphics[width=.06\textwidth]{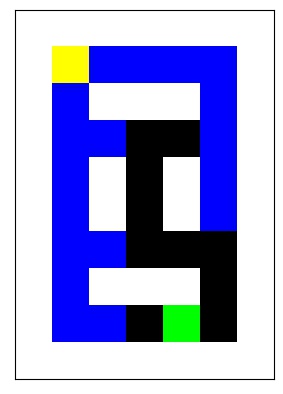} & 
    \includegraphics[width=.06\textwidth]{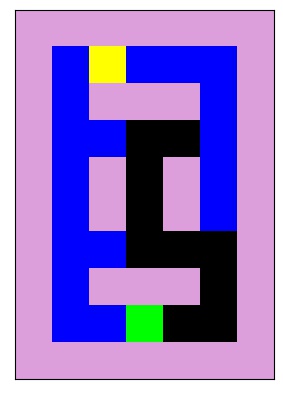}
     \\

   \includegraphics[width=.06\textwidth]{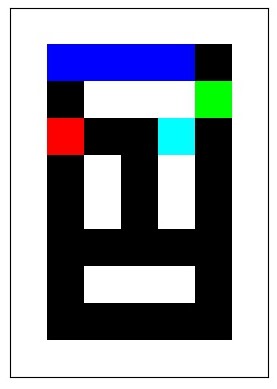} 
    &\includegraphics[width=.06\textwidth]{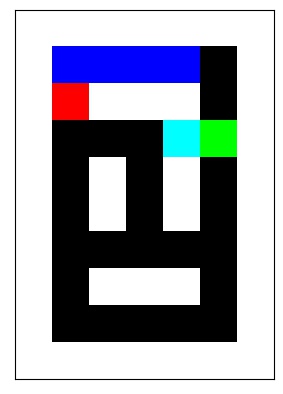} &\includegraphics[width=.06\textwidth]{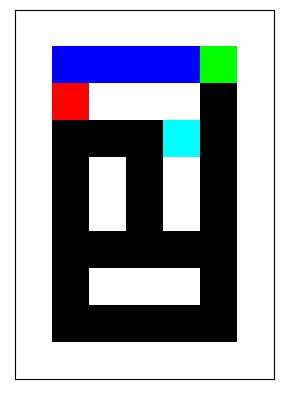} &\includegraphics[width=.06\textwidth]{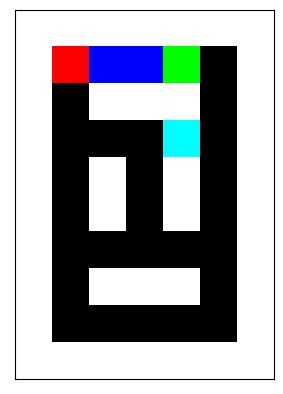} 
    &\includegraphics[width=.06\textwidth]{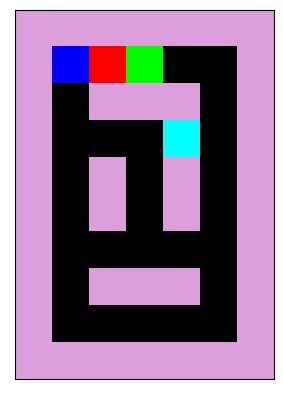} &
   
    \includegraphics[width=.06\textwidth]{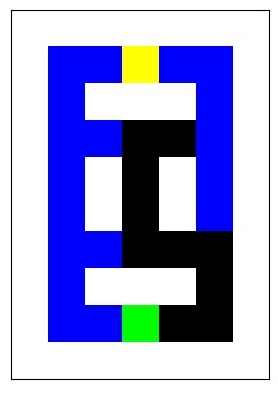} &
    \includegraphics[width=.06\textwidth]{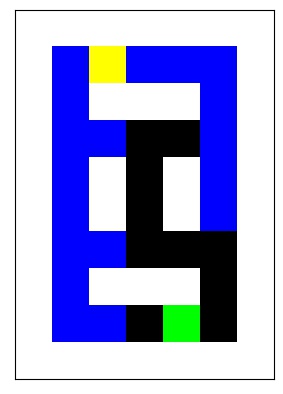} &
    \includegraphics[width=.06\textwidth]{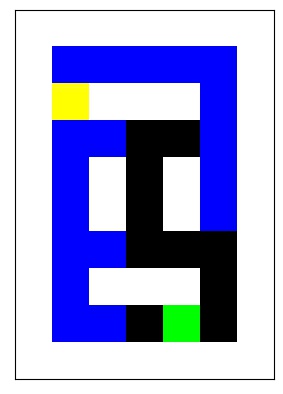} &
    \includegraphics[width=.06\textwidth]{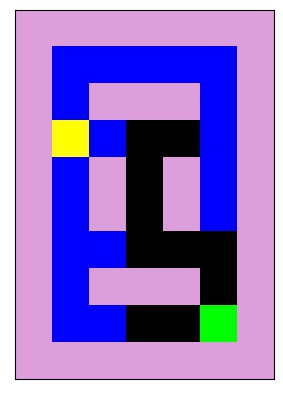} & 

\eat{    
\\
&&&&&
   \includegraphics[width=.06\textwidth]{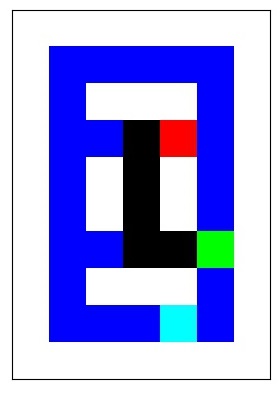} 
    &\includegraphics[width=.06\textwidth]{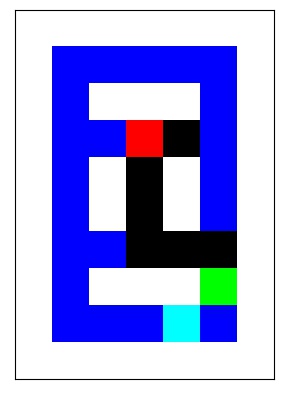}
    &\includegraphics[width=.06\textwidth]{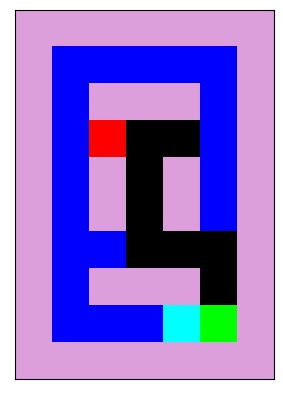} & &
}
 \\\\

\multicolumn{5}{c|}{EAT Scenario 3} &\multicolumn{5}{c}{HUNT Scenario 3}\\ 
   \includegraphics[width=.06\textwidth]{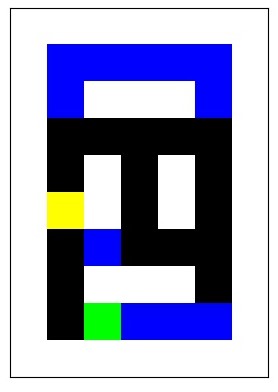} & 
   \includegraphics[width=.06\textwidth]{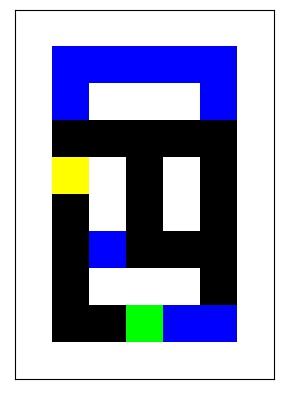} & 
   \includegraphics[width=.06\textwidth]{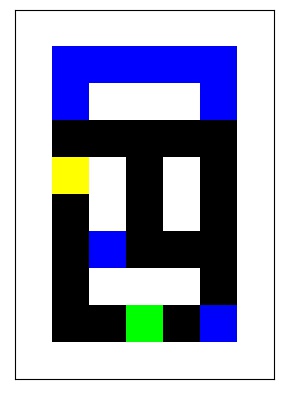} & 
   \includegraphics[width=.06\textwidth]{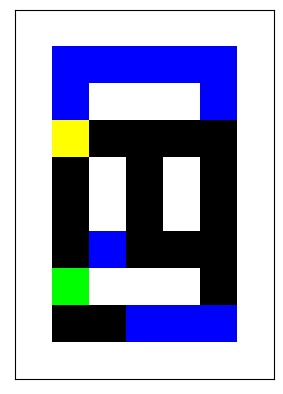} & 
   \includegraphics[width=.06\textwidth]{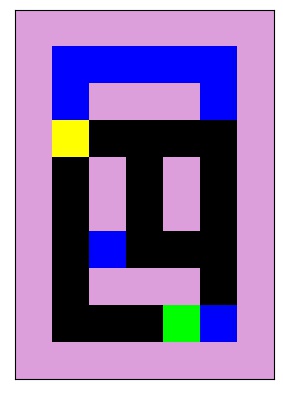} & 
   \includegraphics[width=.06\textwidth]{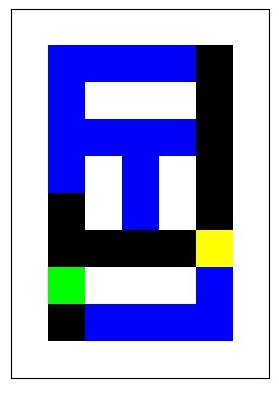} & 
    \includegraphics[width=.06\textwidth]{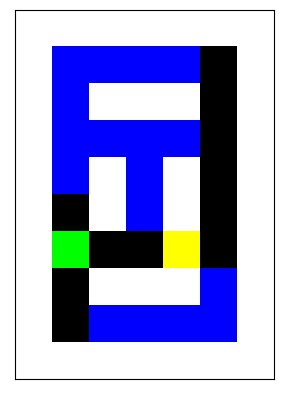} & 
    \includegraphics[width=.06\textwidth]{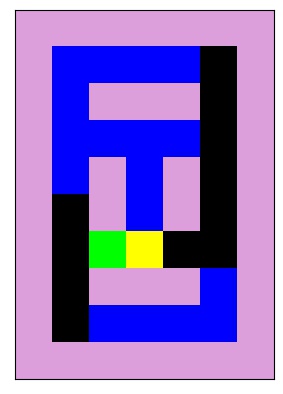} & 
&
\\
   \includegraphics[width=.06\textwidth]{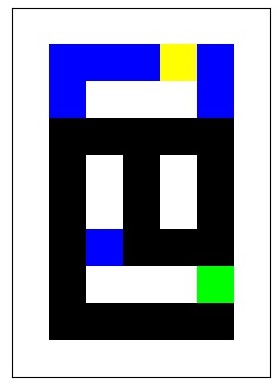} & 
   \includegraphics[width=.06\textwidth]{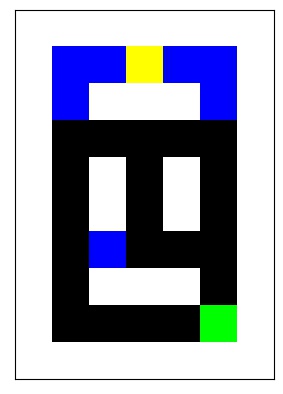} & 
   \includegraphics[width=.06\textwidth]{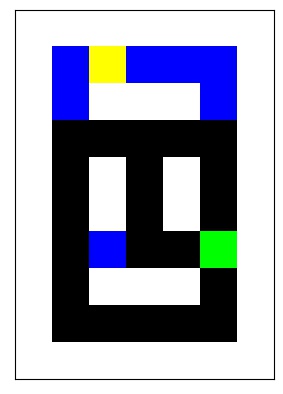} & 
   \includegraphics[width=.06\textwidth]{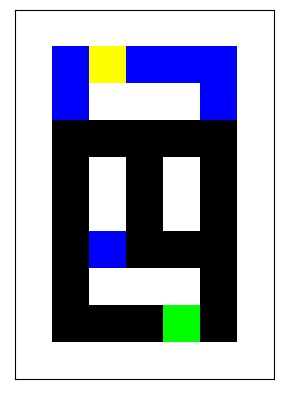} & 
   \includegraphics[width=.06\textwidth]{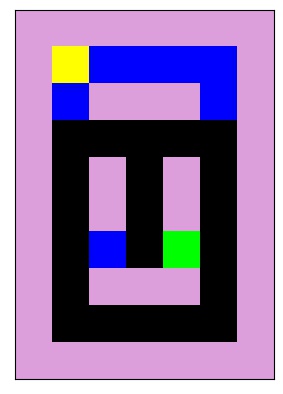} & 
   \includegraphics[width=.06\textwidth]{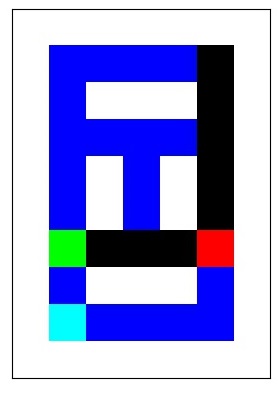} & 
    \includegraphics[width=.06\textwidth]{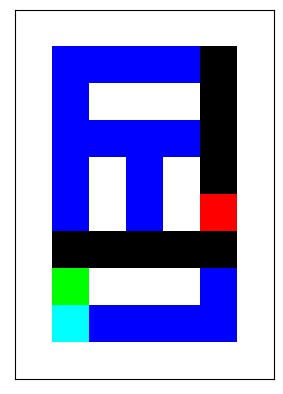} & 
    \includegraphics[width=.06\textwidth]{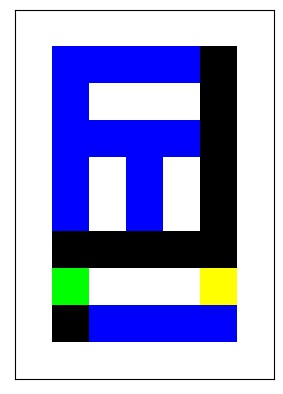} & 
    \includegraphics[width=.06\textwidth]{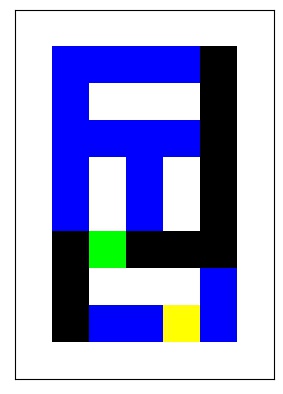} & 
\includegraphics[width=.06\textwidth]{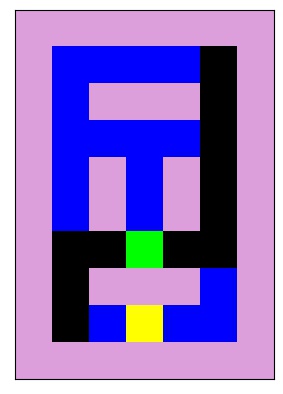} 

    \end{tabular}
  \caption{Illustration of our SSX method on two additional scenarios of minipacman. Two policies, EAT and HUNT, are displayed across the two scenarios. For each scenario, two clusters, one per row, are shown as part of the SSX result. The last board with pink background is a strategic state for each cluster. The color scheme is as follows: green = pacman, red = ghost, yellow = edible ghost, cyan = pill, blue = food, black = food eaten, white/pink=wall. In EAT scenarios, pacman generally ignores the pill and stays away from the ghost (even if the pill has been eaten). In HUNT, pacman generally looks for the pill (but stays away if the ghost is near it) and moves toward the ghost (if the pill has been eaten).}
    \label{fig:minipacman_strategic_states_appendix}
\end{center}
\end{figure*}
Figure \ref{fig:minipacman_strategic_states_appendix} shows results of SSX applied to two additional scenarios for the minipacman example. EAT Scenario 2 shows pacman willing to take a chance of being eaten in order to get more food and EAT Scenario 3 shows that, even though pacman already ate the pill (the ghost is yellow when the pill is eaten), pacman prefers to eat more food rather than head for the ghost. These strategic states contrast directly with those in the HUNT scenarios. Strategic states in Hunt Scenarios 2 and 3 also show pacman eating the pill in order to hunt the ghost rather than eating more food.

\section{Additional User Study Information}
Figure \ref{fig:userstudy_likert} shows the results of qualitative questions (``Was it complete/sufficient/satisfactory/easy to understand?") for both SSX and Viper-D which users rate on a 5-point scale ranging from ``Not at all" to ``Yes absolutely". SSX scores higher across all metrics and differences with Viper-D are statistically significant. These results are consistent with the very different percentage of Unclear selections for SSX and Viper-D, i.e., users found very few SSX explanations to be unclear and hence also scored SSX higher in the qualitative metrics.

\begin{figure*}
\centering
\begin{tabular}{c}
   \includegraphics[width=0.65\textwidth]{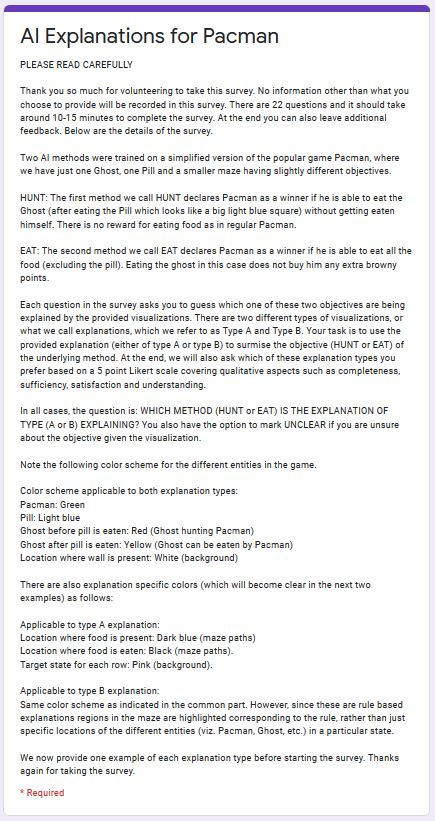}
\end{tabular}
  \caption{Screenshot of user study instructions. SSX explanations are anonymized as Type A explanations and Viper-D explanations are anonymized as Type B explanations.}
    \label{fig:survey_instructions}
\end{figure*}

\begin{figure*}
\centering
\begin{tabular}{c}
   \includegraphics[width=0.9\textwidth]{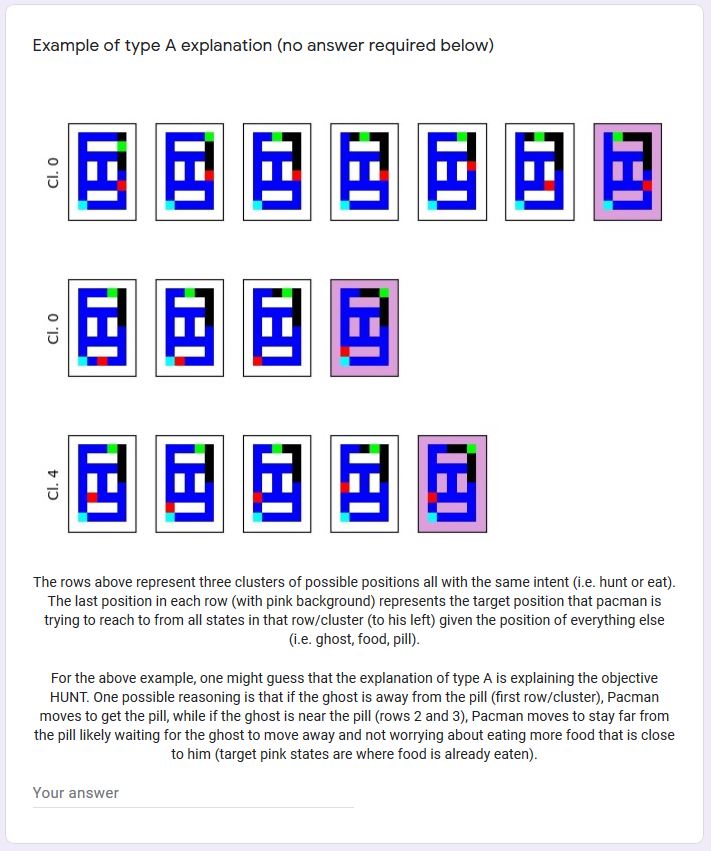}
\end{tabular}
  \caption{Screenshot of SSX explanation example used to train the participant taken from user study. SSX explanations are anonymized as Type A explanations.}
    \label{fig:survey_ssx_example}
\end{figure*}

\begin{figure*}
\centering
\begin{tabular}{c}
   \includegraphics[width=0.85\textwidth]{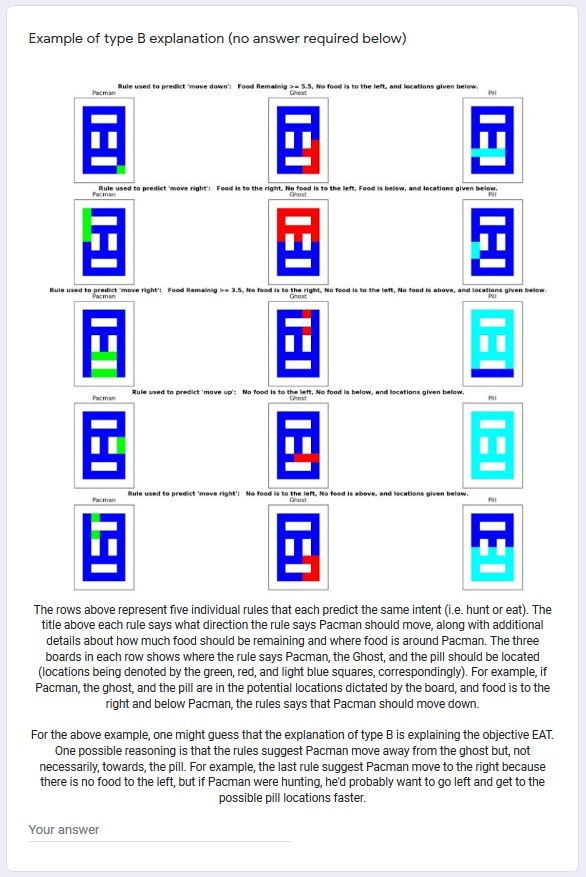}
\end{tabular}
  \caption{Screenshot of Viper-D explanation example used to train the participant taken from user study. Viper-D explanations are anonymized as Type B explanations.}
    \label{fig:survey_viperd_example}
\end{figure*}

\begin{figure*}
\centering
\begin{tabular}{c}
   \includegraphics[width=0.9\textwidth]{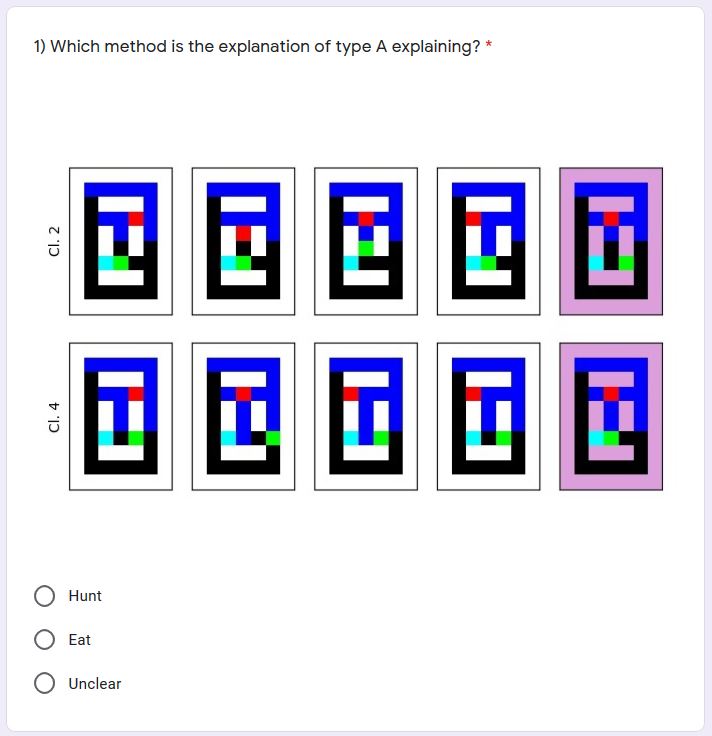}
\end{tabular}
  \caption{Screenshot of SSX explanation survey question taken from user study. SSX explanations are anonymized as Type A explanations.}
    \label{fig:survey_ssx_survey}
\end{figure*}

\begin{figure*}
\centering
\begin{tabular}{c}
   \includegraphics[width=0.9\textwidth]{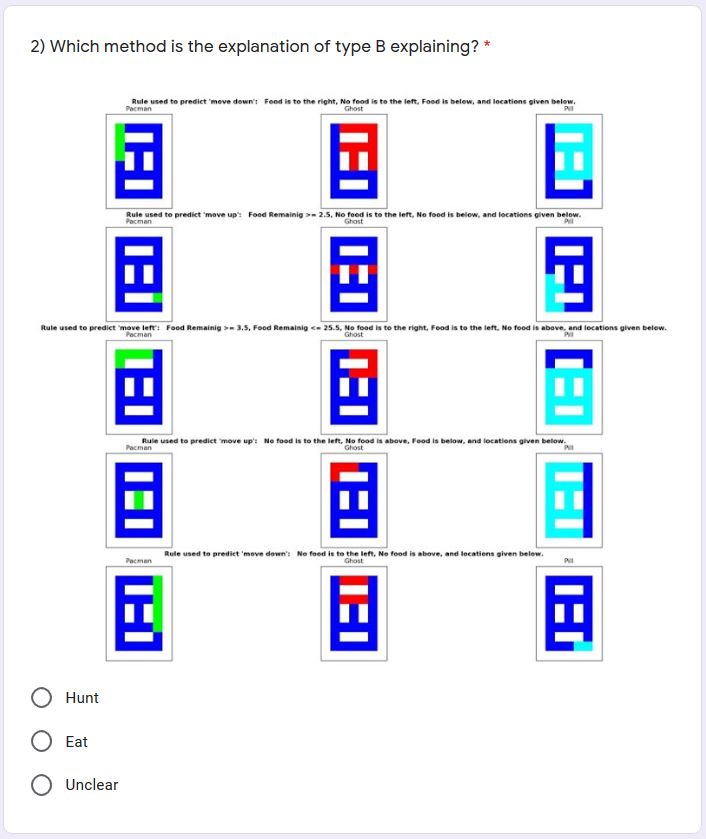}
\end{tabular}
  \caption{Screenshot of Viper-D explanation survey question taken from user study. Viper-D explanations are anonymized as Type B explanations.}
    \label{fig:survey_viperd_survey}
\end{figure*}

Participants in the user study first read a set of instructions where the two environments, hunt and eat, are explained. Further, participants are detailed the color scheme and what they are expected to do with each question of the survey. The full set of instructions are given in Figure \ref{fig:survey_instructions}. Participants were then first trained on one example each with SSX and Viper-D explanations and were given the reasoning one might use in making their choices. The respective training examples are given in Figures \ref{fig:survey_ssx_example} and \ref{fig:survey_viperd_example}. Following the training examples, participants went through a series of 20 questions, 10 from each of SSX and Viper-D explanations. The distribution of correct answers for eat versus hunt is 50/50. An example of each type of question with the choices, Eat, Hunt, and Unclear, are shown in Figures \ref{fig:survey_ssx_survey} and \ref{fig:survey_viperd_survey}.

\eat{\section{Limitations}

Ours being a posthoc explainability method, there is a chance that the insight provided may not be 100\% faithful to the behavior of the expert policy. The meta-state and strategic state estimations, while based on sound approximation schemes, are still approximations and better solutions maybe possible. This gap between the explanation and the true behavior, however, is prevalent even for other posthoc methods not limited to the context of RL (viz. in supervised learning). The current proposal has mainly been tested on discrete state spaces, however, application to continuous spaces in a seamless manner without resorting to discretization would be interesting to look at in the future.}
\end{document}